\documentclass[preprint,12pt,3p]{elsarticle}
\usepackage[utf8]{inputenc}
\usepackage{multirow}
\usepackage{array}
\usepackage[lofdepth,lotdepth]{subfig}
\usepackage[ruled]{algorithm2e}
\usepackage{times,amsmath,amssymb}
\usepackage[T1]{fontenc}
\usepackage[polutonikogreek,english]{babel}
\usepackage{float}
 \usepackage[table,xcdraw]{xcolor}

\usepackage{amssymb}
\journal{neurocomputing}

\begin{document}

\begin{frontmatter}

\title{Ellipsoidal Subspace Support Vector Data Description}

\address[label1]{Faculty of Information Technology and Communication Sciences, Tampere University, FI-33720 Tampere, Finland}
\address[label2]{Department of Engineering, Electrical and Computer Engineering, Aarhus University, DK-8200 Aarhus, Denmark}
\address[label3]{Finnish Environment Institute,
Jyväskylä office, Survontie 9A, FI-40500 Jyväskylä, Finland}

\author[label1]{Fahad Sohrab\corref{cor1}}
\cortext[cor1]{corresponding author}
\ead{fahad.sohrab@tuni.fi}
\author[label3]{Jenni Raitoharju}
\ead{jenni.raitoharju@tuni.fi}
\author[label2]{Alexandros Iosifidis}
\ead{ai@eng.au.dk}
\author[label1]{Moncef Gabbouj}
\ead{moncef.gabbouj@tuni.fi}

\begin{abstract}
In this paper, we propose a novel method for transforming data into a low-dimensional space optimized for one-class classification. The proposed method iteratively transforms data into a new subspace optimized for ellipsoidal encapsulation of target class data. We provide both linear and non-linear formulations for the proposed method. The method takes into account the covariance of the data in the subspace; hence, it yields a more generalized solution as compared to Subspace Support Vector Data Description for a hypersphere. We propose different regularization terms expressing the class variance in the projected space. We compare the results with classic and recently proposed one-class classification methods and achieve better results in the majority of cases. The proposed method is also noticed to converge much faster than recently proposed Subspace Support Vector Data Description.
\end{abstract}

\begin{keyword}
ellipsoidal data description\sep anomaly detection\sep machine learning\sep one-class classification\sep subspace learning
\end{keyword}

\end{frontmatter}

% \linenumbers


\section{Introduction}\label{intro}

The ability of machines to make a concise description of information requires learning from previous experience. Researchers have been trying to develop techniques for accurately modeling data using supervised and unsupervised learning techniques for many decades. In unsupervised learning techniques, patterns are found without any knowledge of class labels \cite{hastie2009unsupervised}. In supervised learning, labeled training data is used to train models for classifying future instances into different categories \cite{kotsiantis2007supervised}. A typical multi-class classification task can be decomposed into several binary class classification tasks, where the aim is to decide to which of the two considered classes samples belong to \cite{fukunaga2013introduction}. In binary classification, the data from both classes are used to train a model. One-class classification is conceptually close to binary classification, but the models for classifying future instances are trained using data from the objects of one particular target class. The aim in one-class classification is to build a model for predicting future instances by using data from a single class only \cite{krawczyk2015one}, \cite{iosifidis2017one}. In practice one-class classification is used when data from one of the classes is scarce.

In one-class classification, the class of interest to be modeled is called target or positive class, while samples from the other unknown class(es) are referred to as outliers or negative samples. Numerous attempts have been made to solve one-class classification tasks \cite{mygdalis2016laplacian}. The three main approaches for solving one-class classification tasks are density based, reconstruction based, and border based methods \cite{tax2002one}. In the density based approach, the description of the target class is based on its density \cite{lee2007density}, which is usually estimated by using popular density estimation methods such as Parzen density, Gaussian model, or mixture of Gaussians \cite{fraley2002model}. In reconstruction based approach, some assumptions about the data generating process are made. The underlying function which represents the target class is obtained by fitting a curve over the data by using prior information, such as data clustering characteristics. Self-organizing maps (SOM) \cite{lloyd1982least} and least-squares quantization \cite{kohonen1998self} are classic examples of reconstruction methods. In border based approaches, a model is created by defining a closed boundary around the target class without estimating its density. One-class Support Vector Machine (OC-SVM) \cite{scholkopfu1999sv} and Support Vector Data Description (SVDD) \cite{tax2004support} are among the popular boundary techniques for one-class classification. In OC-SVM, a hyperplane separating the target class is constructed so that the distance of the hyperplane from the origin is maximized. In SVDD, a hypersphere is formed around the target class data by minimizing the volume of hypersphere in a given feature space.  

SVDD has been justified over time  as a powerful data description method and it has been used in many different application domains for solving one-class classification problems. For example, in \cite{zhao2010identification}, SVDD is found to be an excellent choice for solving the problem of identification of freshness of eggs using near infrared spectroscopy (NIR) with an imbalanced number of training samples. In \cite{lee2018terrain}, a terrain classification method for ensuring navigation safety of mobile service robots based on SVDD is proposed. To enhance the performance of SVDD, numerous extensions and hybridization techniques have been proposed \cite{lee2007density}, \cite{mygdalis2016graph}, \cite{wang2019dynamic}, \cite{kenaza2018efficient}, \cite{mygdalis2018semi} \cite{rahmanimanesh2019adaptive}. The main extensions of SVDD can be categorized into four main categories. In the first category of extensions, the techniques are focused on manipulating the structure of data, such as associating a confidence coefficient with all training instances which deals with the uncertainty of data \cite{liu2013svdd}. In the second category, the performance is enhanced by proposing new non-linear methods and reducing the complexity of algorithms \cite{banerjee2007fast}, \cite{liu2010fast}. Techniques for handling non-stationary data in the context of one-class classification falls in the third category of extensions \cite{camci2008general}. In the fourth category, different changes are proposed in the shape of the boundary encapsulating the target data \cite{forghani2011support}.

A popular alternative to the spherical SVDD is Ellipsoidal SVDD (E-SVDD) \cite{forghani2011support}, \cite{ghasemigol2009ellipse}. E-SVDD forms a unique hyperellipsoid with a minimum volume covering most of the target data. An ellipsoid, unlike a hypersphere, takes into account the difference in variance for each dimension as well as covariance between them. A hypersphere, characterized only by a radius and a center will result in superfluous regions which do not contain any target objects in the input space \cite{wang2017ellipsoidal}. Ellipsoids with a minimum volume containing the target data have applications spanning over many different fields. For example, in \cite{ghasemigol2010intrusion}, it is used to detect intrusion in computer networks and, in \cite{rimon1997obstacle}, it is used to estimate the distance between a robot and its surrounding environment for obstacle collision avoidance. An ellipsoid is preferred for heterogeneous data in the input space because its shape is less conservative than a sphere. However, there are some difficulties in kernelizing the algorithms. The kernel trick cannot be applied directly to E-SVDD because its formulation includes outer products rather than inner products \cite{wang2016ellipsoidal}.

In this paper, we propose a novel subspace learning algorithm for ellipsoidal one-class classification. The proposed method takes into account the covariance of data in the subspace so that the boundary created around the target class is a better fit. The proposed method finds a projection along with data description iteratively by minimizing the volume of the hyperellipsoid. We also propose a non-linear version of the algorithm by exploiting the non-linear projection trick (NPT) \cite{kwak2013nonlinear}. The proposed method is called Ellipsoidal Subspace Support Vector Data Description (ES-SVDD), since it is analogous to Subspace Support Vector Data Description (S-SVDD) proposed in \cite{sohrab2018subspace} but offers more flexibility by using hyperellipsoids instead of hyperspheres.

The rest of the paper is organized as follows. In Section \ref{ess}, we present an overview of related works. In Section \ref{ess_gesvdd}, a detailed derivation of the newly proposed method is presented. In Section \ref{experiments}, we provide and discuss the experimental protocol along with the obtained results and, finally, conclusions are drawn in Section \ref{conculsion}.

\section{Background and related work}\label{ess}
One-class classification has been studied extensively in recent years and the approaches predominantly focus on data description in a given feature space \cite{tax2002one}, \cite{tax2004support}, \cite{liu2013svdd}. On the other hand, feature selection and subspace learning have been an active research area in machine learning, primarily for challenges with data available for all categories \cite{cai2018feature}, \cite{ji2010shared}. The aim is to avoid the curse of dimensionality in the original feature space by modeling the given data in a lower dimensional space.

In feature selection methods, a subset of representative features is selected by following some criterion \cite{guyon2003introduction}, \cite{zhao2011similarity}, \cite{zhou2016global}. The two main approaches for feature selection are the \textit{filter} approach and the \textit{wrappers} approach. In the filter approaches, the main focus is on the intrinsic characteristics of the data and they do not take into account any classification algorithm. On the other hand, the wrappers approaches are dependent only on a specific classification algorithm \cite{kermani2019global}. 

In subspace learning, the features are transformed from original feature space to a lower-dimensional subspace \cite{gu2011joint}. Most of the existing subspace learning methods, particularly for anomaly detection, follow three general steps \cite{bacher2016subspace}, \cite{nguyen20134s}: First, the features are selected randomly by applying random projections to the attributes. Second, classical algorithms are applied locally in each subspace and scores (e.g., voting) are computed. Finally, all the scores are aggregated to compute a global score for classification.

The focus of our paper is to find an optimized subspace for one-class classification. We review the classical one-class classification method, SVDD, in Section \ref{sectionSVDD} and also provide an overview of S-SVDD and graph embedded one-class classifiers in Sections \ref{sectionSSVDD} and \ref{geocc}, respectively. 
\subsection{Support Vector Data Description}\label{sectionSVDD}
Let us denote the data points to be enclosed inside a closed boundary by a matrix $\mathbf{X}=[\mathbf{x}_{1},\mathbf{x}_{2},\dots \mathbf{x}_{N}],\mathbf{x}_{i} \in \mathbb{R}^{D}$, where $N$ is total number of instances and $D$ is dimensionality of data in the original feature space. All the data samples represented by $\mathbf{X}$ belong to the same class. 

SVDD finds a spherical boundary around the data by minimizing the volume of a hypersphere enclosing all the target class data:
\begin{eqnarray}\label{erfuncSVDD}
\min \quad & F(R,\mathbf{a}) = R^2 \nonumber\\
\textrm{s.t.} \quad & \| {\mathbf{x}_i} - \mathbf{a} \|^2_{2} \le R^2, \:\: \forall i\in\{1,\dots,N\},
\end{eqnarray}  
where $R$ is the radius of hypersphere and $\mathbf{a}\in\mathbb{R}^{D}$ is the center of the hypersphere in the given feature space. Slack variables $\xi_i,\:i=1,\dots,N$ are introduced for allowing the possibility of data points being outliers, hence the optimization problem changes to
\begin{eqnarray}\label{erfuncSVDD2}
\min \quad & F(R,\mathbf{a}) = R^2 + C\sum_{i=1}^{N} \xi_i \nonumber\\
\textrm{s.t.} \quad & \|\mathbf{x}_i - \mathbf{a}\|_2^2 \le R^2 + \xi _i,\nonumber\\
&\xi_i \ge 0, \:\: \forall i\in\{1,\dots,N\},
\end{eqnarray}  
where $C>0$ is a hyperparameter which controls the trade-off between the volume of the sphere and the amount of data outside the sphere. The Lagrangian dual of Eq. \eqref{erfuncSVDD2} reduces to
\begin{eqnarray}\label{langSVDD}
L &=& \sum_{i=1}^{N} \alpha_i \mathbf{x}_i^{\intercal} \mathbf{x}_i - \sum_{i}^{N}\sum_{j}^{N} \alpha_i \alpha_j \mathbf{x}_i^{\intercal} \mathbf{x}_j,
\end{eqnarray}
subject to $0 \le \alpha_i \le C$. Maximizing Eq. \eqref{langSVDD} gives a set of $\alpha_i$ for corresponding data points. The samples with $\alpha_i > 0$ are the support vectors defining the data description \cite{tax2004support}. The samples corresponding to $0 < \alpha_i < C$ lie on the boundary
of the hypersphere and those with $\alpha_i = C$ are outliers.

\subsection{Subspace Support Vector Data Description}\label{sectionSSVDD}
In S-SVDD \cite{sohrab2018subspace}, a projection matrix $\mathbf{Q}$ is determined to map data from the original space $\mathbb{R}^{D}$ to a new optimized lower dimensional space $\mathbb{R}^{d},\:d<D$, so that the data is more suitable for one-class classification:
\begin{eqnarray}\label{ssvdd}
\min \quad & F(R,\mathbf{a}) = R^2 + C\sum_{i=1}^{N} \xi_i \nonumber\\
\textrm{s.t.} \quad & \|\mathbf{Qx}_i - \mathbf{a}\|_2^2 \le R^2 + \xi _i,\nonumber\\
&\xi_i \ge 0, \:\: \forall i\in\{1,\dots,N\},
\end{eqnarray}  
where $\mathbf{a}\in\mathbb{R}^{d}$ is the center of the hypersphere in lower \textit{d}-dimensional space. The method iteratively solves the SVDD in the current subspace to obtain the data description parameters $\alpha_i, \:i=1,\dots,N$, and then updates the subspace projection by optimizing an augmented version of the Lagrangian:
\begin{equation}\label{langSSVDD}
L= \sum_{i=1}^{N} \alpha _i  \mathbf{x}_i^{\intercal} \mathbf{Q}^{\intercal} \mathbf{Q} \mathbf{x}_i - \sum_{i=1}^{N}\sum_{j=1}^{N} \alpha_i \mathbf{x}_i^{\intercal} \mathbf{Q}^{\intercal} \mathbf{Q} \mathbf{x}_j \alpha_j + \beta\psi,
\end{equation}  
where $\psi$ is a regularization term expressing the class variance in the low dimensional space and $\beta$ is a regularization parameter controlling the importance of the $\psi$, where
\begin{equation}
\label{generalconstraintpsi} 
\psi = \text{Tr}(\mathbf{Q}\mathbf{X} \boldsymbol{\lambda}\boldsymbol{\lambda}^\intercal \mathbf{X}^\intercal\mathbf{Q}^\intercal),
\end{equation}  
where $\text{Tr}(.)$ is the trace operator and $\boldsymbol{\lambda} \in \mathbb{R}^{N}$ is a vector controlling the contribution of each training sample. $\mathbf{Q}$ is updated by using the gradient of Eq. \eqref{langSSVDD}, i.e.,
\begin{eqnarray}\label{Quopdate}
\mathbf{Q}\leftarrow \mathbf{Q}-\eta\Delta L,
\end{eqnarray}
where $\eta$ is the learning rate. A non-linear version of S-SVDD employing the kernel trick is also proposed in \cite{sohrab2018subspace}. 

\subsection{Graph embedded one-class classifiers}\label{geocc}
Graph embedded one-class classifiers constitute extensions of the OC-SVM and SVDD by incorporating generic graph structures in their optimization process. The generic graph structures express geometric data relationships of the target class in the data. For example, Graph Embedded SVDD (GE-SVDD) \cite{mygdalis2016graph} optimization problem is formulated as

\begin{eqnarray}\label{GESVDD}
\min \quad & F(R,\mathbf{a}) = R^2 + C\sum_{i=1}^{N} \xi_i \nonumber\\
\textrm{s.t.} \quad & \big(\phi (\mathbf{x}_i)-\mathbf{a}\big)^\intercal\mathbf{S}^{-1}\big(\phi (\mathbf{x}_i)-\mathbf{a}\big) \le R^2 + \xi_i, \nonumber\\
&\xi_i \ge 0, \forall i\in \{1,\dots,N\},
\end{eqnarray}
where $\phi(.)$ is any non-linear function used for mapping the training samples from the input feature space to the kernel space. The matrix $\mathbf{S}$ contains the geometric data relationships. For example, in PCA, the scatter of training data can be expressed as
\begin{eqnarray}\label{matrixS}
\mathbf{S}=\frac{1}{N}\mathbf{\Phi}\Big(\mathbf{I}-\frac{1}{N}\mathbf{1}\mathbf{1}^\intercal\Big) \mathbf{\Phi}^\intercal =\mathbf{\Phi} \mathbf{L} \mathbf{\Phi}^\intercal,
\end{eqnarray}
where $\mathbf{1} \in \mathbb{R}^N$ is a vector containing all values as ones, $\mathbf{I}\in\mathbb{R}^{N\times N} $ is an identity matrix, and $\mathbf{\Phi}$ is a matrix that contains the training data representations in kernel space.

The Lagrangian of GE-SVDD is
\begin{eqnarray}\label{GEsvddLang}
L= \sum_{i=1}^{N} \alpha _i  \phi (\mathbf{x}_i)^{\intercal} \mathbf{S}^{-1}  \phi (\mathbf{x}_i)- \sum_{i=1}^{N}\sum_{j=1}^{N} \alpha_i \phi (\mathbf{x}_i)^{\intercal} \mathbf{S}^{-1}  \phi (\mathbf{x}_j) \alpha_j
\end{eqnarray}
It has been shown in \cite{mygdalis2016graph} that the optimization problem in Eq. \eqref{GEsvddLang} is equivalent to the problem of SVDD in a transformed feature space.

\section{Ellipsoid Subspace Support Vector Data Description}\label{ess_gesvdd}
Our aim is to find a projection matrix  $\mathbf{Q} \in \mathbb{R}^{d \times D}$ to be used for transforming the data to an optimized subspace suitable for one-class classification. In the following analysis, we assume that the data has been centered by setting $\mathbf{X}\leftarrow \mathbf{X}-\boldsymbol{\mu}$, where $\boldsymbol{\mu}$ is the mean of the given training data. The mapping from the original feature space with dimensionality $D$ to a subspace with dimensionality $d\le D $ is carried out. The mapping is done to transform the data so that it is more suitable to be encapsulated inside an ellipsoid with a minimum volume. 

The optimization problem is formulated as
\begin{eqnarray}\label{primalobj_gesvdd}
\min \quad & F(R,\mathbf{a}) = R^2 + C\sum_{i=1}^{N} \xi_i \nonumber\\
\textrm{s.t.} \quad & (\mathbf{Qx}_i-\mathbf{a})^\intercal\mathbf{E}^{-1}(\mathbf{Qx}_i-\mathbf{a}) \le R^2 + \xi_i, \nonumber\\
&\xi_i \ge 0, \forall i\in \{1,\dots,N\},
\end{eqnarray}
where $\mathbf{a}$ is the center of the hyperellipsoid and $\mathbf{E}=\mathbf{QX}\mathbf{X}^{\intercal} \mathbf{Q}^{\intercal}$ is the covariance matrix of the data in \textit{d}-dimensional space. The inverse of covariance matrix $\mathbf{E}$, also known as the concentration or precision matrix is symmetric and positive definite  $\mathbf{E}^{-1}\in  \mathbb{R}^{d\times d}$. By defining a new vector $\mathbf{u}=\mathbf{E}^{-\frac{1}{2}}\mathbf{a}$, Eq. \eqref{primalobj_gesvdd} can be written as
\begin{eqnarray}\label{primalobj_gesvdd2}
\min \quad & F(R,\mathbf{u}) = R^2 + C\sum_{i=1}^{N} \xi_i \nonumber\\
\textrm{s.t.} \quad &  \| \mathbf{E}^{-\frac{1}{2}}\mathbf{Qx}_i-\mathbf{u} \|^2_{2} \le R^2 + \xi_i, \nonumber\\
&\xi_i \ge 0, \forall i\in \{1,\dots,N\}.
\end{eqnarray}
The data in the subspace is represented by
\begin{equation}\label{subspacedata}
\mathbf{y}_i = \mathbf{Q} \mathbf{x}_i, \:\:i=1,\dots,N.
\end{equation}
The constraints in Eq. \eqref{primalobj_gesvdd2} can be incorporated into its corresponding objective function by using Lagrange multipliers: 
\begin{eqnarray}\label{lagrangian0}
L= R^2+C\sum_{i=1}^{N}\xi_i-\sum_{i=1}^{N}\alpha_i\big(R^2+\xi_i 
-(\mathbf{E}^{-\frac{1}{2}}\mathbf{y}_i)^\intercal
\mathbf{E}^{-\frac{1}{2}}\mathbf{y}_i +
2\mathbf{u}^\intercal\mathbf{E}^{-\frac{1}{2}}\mathbf{y}_i-\mathbf{u}^\intercal\mathbf{u}\big)-\sum_{i=1}^{N}\gamma_i\xi_i
\end{eqnarray}
with Lagrange multipliers $\alpha_i\ge0$ and $\gamma_i\ge0$. 

By setting partial derivatives with respect to $R$, $\mathbf{u}$ and $\xi_i$ to zero, we get
\begin{eqnarray}
\frac{\partial L}{\partial R}=0 &\Rightarrow& \sum_{i=1}^{N} \alpha_i = 1 \label{der1} \\
\frac{\partial L}{\partial {\mathbf{u}}}=0 &\Rightarrow& {\mathbf{u}} = \sum_{i=1}^{N} \alpha_i \mathbf{E}^{-\frac{1}{2}}\mathbf{Qx}_i \label{der2} \\
\frac{\partial L}{\partial \xi _i}=0 &\Rightarrow& C- \alpha _i - \xi _i  = 0. \label{der3}
\end{eqnarray} 
By substituting Eqs. \eqref{der1}-\eqref{der3} into Eq. \eqref{lagrangian0} we get
\begin{eqnarray}\label{LL}
L = \sum_{i=1}^{N} \alpha _i {\mathbf{x}}_i^\intercal {\mathbf{Q}}^\intercal \mathbf{E}^{-1} \mathbf{Qx}_i -  \sum_{i=1}^{N}\sum_{j=1}^{N} \alpha _i \mathbf{x}_i^\intercal \mathbf{Q}^\intercal\mathbf{E}^{-1} \mathbf{Qx}_j \alpha _j.
\end{eqnarray} 
We can use SVDD to solve Eq. \eqref{LL} for getting $\alpha_i$ values. The concentration matrix $\mathbf{E}^{-1}$ is equivalent to
\begin{eqnarray}\label{cov}
\mathbf{E}^{-1}=(\mathbf{Q}\mathbf{X}\mathbf{X}^\intercal \mathbf{Q}^{\intercal})^{-1}.
\end{eqnarray}
By putting Eq. \eqref{cov} in Eq. \eqref{LL} we get
 \begin{eqnarray}\label{LLcov}
L = \sum_{i=1}^{N} \alpha _i \mathbf{x}_i^\intercal \mathbf{Q}^\intercal (\mathbf{Q}\mathbf{X}\mathbf{X}^\intercal \mathbf{Q}^{\intercal})^{-1} \mathbf{Qx}_i -  \sum_{i=1}^{N}\sum_{j=1}^{N} \alpha _i \mathbf{x}_i^\intercal \mathbf{Q}^\intercal(\mathbf{Q}\mathbf{X}\mathbf{X}^\intercal \mathbf{Q}^{\intercal})^{-1} \mathbf{Qx}_j \alpha _j.
\end{eqnarray} 

We add an extra term $\Upsilon$ to Eq. \eqref{LLcov} as a regularization term expressing the class variance in the projected space, also taking into account the concentration matrix. Hence, Eq. \eqref{LLcov} now becomes
 \begin{eqnarray}\label{LLcov_phi}
L = \sum_{i=1}^{N} \alpha _i \mathbf{x}_i^\intercal \mathbf{Q}^\intercal (\mathbf{Q}\mathbf{X}\mathbf{X}^\intercal \mathbf{Q}^{\intercal})^{-1} \mathbf{Qx}_i -  \sum_{i=1}^{N}\sum_{j=1}^{N} \alpha _i \mathbf{x}_i^\intercal \mathbf{Q}^\intercal(\mathbf{Q}\mathbf{X}\mathbf{X}^\intercal \mathbf{Q}^{\intercal})^{-1} \mathbf{Qx}_j \alpha _j + \beta\Upsilon,
\end{eqnarray} 
where $\beta$ controls the importance of regularization term and is used as a hyperparameter. $\Upsilon$ is defined as follows:
\begin{eqnarray}\label{Reg1}
\Upsilon=\text{Tr}(\mathbf{E}^{-\frac{1}{2}}\mathbf{QX}\boldsymbol{\lambda}\boldsymbol{\lambda}^\intercal\mathbf{X}^\intercal\mathbf{Q}^\intercal\mathbf{E}^{-\frac{\intercal}{2}}),
\end{eqnarray} 
where $\boldsymbol{\lambda}$ can take three different forms. In the first form, all elements in $\boldsymbol{\lambda}$ take the value of 1 and, hence, all the samples are used to describe the covariance of the class. In the second form, $\boldsymbol{\lambda}$ is replaced by $\boldsymbol{\alpha}$, which means that the samples belonging to the boundary and outside the boundary are used to describe the covariance of the class. In the third form, the $\lambda_i$ values are replaced by ${\alpha_i}$ values of the samples belonging to the boundary and zero for other instances. The first, second and third forms of the regularization terms are expressed as $\Upsilon_1$, $\Upsilon_2$, and $\Upsilon_3$ hereinafter.

In our experiments, we also consider the regularization term expressing the class variance in the projected space without taking into account the concentration matrix. This is achieved by replacing the covariance matrix $\mathbf{E}$ with the identity matrix $\mathbf{I}$ in Eq. \eqref{Reg1}. By doing so, the regularization term $\Upsilon$ becomes equivalent to $\psi$ as described in Eq. \eqref{generalconstraintpsi}. Analogous to regularization term $\Upsilon$, $\psi$ can also take different forms by changing $\boldsymbol{\lambda}$ and similarly hereinafter we refer to all those cases by $\psi_1$, $\psi_2$ and $\psi_3$. The methods used with $\psi$ and $\Upsilon$ are denoted by ES-SVDD$\psi_m$ and ES-SVDD$\Upsilon_m$ ($m=1,2,3$), respectively. We refer to the case, where no regularization term is used in ES-SVDD, as ES-SVDD$\psi_0\Upsilon_0$.

Eq. \eqref{LLcov_phi} can be further simplified and written as 
\begin{eqnarray}\label{traceform}
L=\text{Tr}((\mathbf{QX}\mathbf{X}^{\intercal}\mathbf{Q}^{\intercal})^{-1}\mathbf{QX}(\mathbb{A}-\boldsymbol{\alpha} \boldsymbol{\alpha}^\intercal)\mathbf{X}^\intercal\mathbf{Q}^{\intercal})+\beta\Upsilon,
\end{eqnarray}  
where $\mathbb{A}$ is a diagonal matrix having $\alpha_i$ values in its diagonal and $\boldsymbol{\alpha}$ is a vector of $\alpha_i$'s. We use gradient of Eq. \eqref{traceform} to update the projection matrix. The gradient can be solved using identity 126 in \cite{IMM2012-03274}:
\begin{eqnarray}\label{deltaL}
\Delta L=  2\mathbf{E}^{-1}\mathbf{QX}(\mathbb{A}-\boldsymbol{\alpha} \boldsymbol{\alpha}^\intercal)\mathbf{X}^\intercal - 2\mathbf{E}^{-1}\mathbf{QX}(\mathbb{A}-\boldsymbol{\alpha} \boldsymbol{\alpha}^\intercal)\mathbf{X}^\intercal    \mathbf{Q} ^\intercal  \mathbf{E}^{-1}  \mathbf{Q}  \mathbf{X}  \mathbf{X}^\intercal  +\beta \Delta\Upsilon,
\end{eqnarray}  
where
\begin{eqnarray}\label{deltaphi}
\Delta\Upsilon= 2\textbf{E}^{-1}\mathbf{Q}\mathbf{X}\boldsymbol{\lambda}\boldsymbol{\lambda}^\intercal \mathbf{X}^\intercal -2\textbf{E}^{-1}\mathbf{QX}\boldsymbol{\lambda}\boldsymbol{\lambda}^\intercal \mathbf{X}^\intercal\mathbf{Q}^\intercal \textbf{E}^{-1}\mathbf{Q}\mathbf{X}\mathbf{X}^\intercal.
\end{eqnarray} 
When $\psi$ is used as a regularization term, we use $\Delta\psi$ instead of $\Delta\Upsilon$ in Eq. \eqref{deltaL}:
\begin{eqnarray}
\label{deltapsi}
\Delta \psi=2\mathbf{Q}\mathbf{X}\boldsymbol{\lambda} \boldsymbol{\lambda}^\intercal\mathbf{X}^\intercal.
\end{eqnarray}  
We obtain an optimised data projection matrix along with optimised data description in a two-step iterative process. In first step, the $\alpha_i$ values are computed by maximizing Eq. \eqref{LL}. In second step, $\mathbf{Q}$ is updated through the gradient descent after computing the gradient by using Eq. \eqref{traceform}. In order to obtain an orthogonal projection, we impose the orthogonality constraint $\mathbf{Q}\mathbf{Q}^\intercal= \mathbf{I}$. We orthogonalize and normalize $\mathbf{Q}$ during the two-step iterative process. Algorithm 1 presents the whole algorithm.

\begin{algorithm}[ht]
  \caption{Linear ES-SVDD optimization}\label{algo}
\SetAlgoLined

\SetKwInOut{Input}{Input}
\SetKwInOut{Output}{Output}
\Input{$\mathbf{X}, \beta, \eta, d, C, k_{max}$}
 \Output{$\mathbf{Q}$, $R$, $\mbox{\boldmath$\alpha$}$}  
 \vspace{3mm}
Random initialization of $\mathbf{Q}$\ ; \\
  Initialize $k= 1$\;
 \vspace{3mm}
 \While{$k< k_{max}$ }{
    \vspace{3mm}
    Compute concentration matrix $\mathbf{E}^{-1}$ using Eq. \eqref{cov} \;
    
    Solve $\alpha_i,\:i=1,\dots,N$ with SVDD using Eq. \eqref{LL}\;
    
    \vspace{3mm}
        Calculate $\Delta L$ using Eq. \eqref{deltaL}\;
    Update $\mathbf{Q} \leftarrow \mathbf{Q} - \eta \Delta L$\;
    
    \vspace{3mm}
    Orthogonalize $\mathbf{Q}$ using QR decomposition\;
    Row normalize $\mathbf{Q}$ using $l_2$ norm\;
   
    % \textbf{Get} $\mathbf{Q}^{*}=\mathbf{Q}^{(k)}$\;
      $k \leftarrow k+1$ 
   }
   \vspace{3mm}
   
   // Data description in the optimized subspace\\
     Compute concentration matrix $\mathbf{E}^{-1}$ using Eq. \eqref{cov}\\
     Calculate $\alpha_i,\:i=1,\dots,N$ with SVDD using Eq. \eqref{LL} \;
% \vspace{5mm}
\end{algorithm}

\subsection{Non-linear Ellipsoidal Subspace Support Vector Data Description}\label{nonlinearessvdd}
The non-linear ellipsoidal subspace SVDD is not trivial, because the kernel trick cannot be applied directly due to the outer products involved in its derivation. To avoid this problem, we follow the NPT based solution described below \cite{kwak2013nonlinear}. We first compute a noncentered kernel matrix $\mathbf{K=\Phi}^{\intercal}\mathbf{\Phi}$ using the radial basis function kernel as

\begin{equation}\label{RBFkernel}
\mathbf{K}_{ij} = \exp  \left( \frac{ -\| \mathbf{x}_{i} - \mathbf{x}_{j}\|_2^2 }{ 2\sigma^2 } \right),
\end{equation} 
where $\sigma$ is a hyperparameter scaling the distance between $\mathbf{x}_i$ and $\mathbf{x}_j$. The kernel matrix is centered as
\begin{eqnarray}\label{centerK}
\mathbf{\Hat{K}} = (\mathbf{I}- \mathbf{J}) \mathbf{K} ( \mathbf{I}-\mathbf{J}),
\end{eqnarray}
where $\mathbf{J}\in\mathbb{R}^{N\times N} $ is a matrix defined as
\begin{eqnarray}\label{ennpt}
\mathbf{J} = \frac{1}{N}\mathbf{1} \mathbf{1}^\intercal.
\end{eqnarray}
The centered kernel matrix is decomposed by using eigendecomposition:
\begin{eqnarray}\label{eigen}
\mathbf{\Hat{K}} = \mathbf{U}\mathbf{A}\mathbf{U}^\intercal, 
\end{eqnarray}
where $\mathbf{A}$ contains the non-negative eigenvalues of the centered kernel matrix in its diagonal and the columns of $\mathbf{U}$ contain the corresponding eigenvectors. Finally, the data in the reduced dimensional kernel space is obtained as
\begin{eqnarray}\label{nptdata}
\mathbf{\Phi} = (\mathbf{A}^{\frac{1}{2}})^{+} \mathbf{U}^{+} {\mathbf{\Hat{K}} },
\end{eqnarray}
where $+$ sign in the superscript denotes the pseudo-inverse.

After applying NPT, we continue by considering $\mathbf{\Phi}$ as our input data. This allows as to use the linear E-SVDD formulation to obtain a non-linear transformation.
\subsection{Test phase}
During the test phase of the linear case, a test instance $\mathbf{x}_*$ is first mapped to the optimized lower \textit{d}-dimensional space as
\begin{eqnarray}\label{test}
\mathbf{y}_* = \mathbf{Q} \mathbf{x}_*.
\end{eqnarray}
The decision to classify the instance as target or outlier is taken on the basis of its distance from the center of data description in the \textit{d}-dimensional space. The distance is calculated as follows:
\begin{eqnarray}\label{testdistance}
\|\mathbf{E}^{-\frac{1}{2}}\mathbf{y}_{*} - \mathbf{u}\|_2^2= (\mathbf{E}^{-\frac{1}{2}}\mathbf{y}_{*})^\intercal\mathbf{E}^{-\frac{1}{2}}\mathbf{y}_{*} - 2(\mathbf{E}^{-\frac{1}{2}}\mathbf{y}_{*})^\intercal\mathbf{u} +\mathbf{u}^\intercal\mathbf{u},
\end{eqnarray}
where $\mathbf{u}$ can be solved with Eq. \eqref{der2}. If  $\|\mathbf{y}_{*} - \mathbf{u}\|_2^2 \le R^2$, the test instance is classified as positive, as it will fall inside the boundary of the data description. The test instance is classified as negative if  $\|\mathbf{y}_{*} - \mathbf{u}\|_2^2 > R^2$. The threshold $R^2$ for taking the decision is calculated as follows:
 \begin{eqnarray}\label{rsquare}
R^2= (\mathbf{E}^{-\frac{1}{2}}\mathbf{s})^\intercal\mathbf{E}^{-\frac{1}{2}}\mathbf{s}- 2 \mathbf{u}^\intercal\mathbf{s} 
+ \mathbf{u}^\intercal\mathbf{u},
\end{eqnarray}
where $\mathbf{s}$ is any support vector with $0 < \alpha_i < C$. 

During the test phase for non-linear ES-SVDD, we use NPT by first computing the kernel vector as
\begin{eqnarray}\label{kvector}
\mathbf{k}_{*} = \mathbf{\Phi}^T \phi(\mathbf{x}_{*}).
\end{eqnarray}
The kernel vector is then centered as
\begin{eqnarray}\label{centerKtest}
\mathbf{{\Hat{k}}}_{*}= (\mathbf{I}- \mathbf{J}) [  \mathbf{{{k}}}_{*}-\frac{1}{N}\mathbf{K} \mathbf{1}].
\end{eqnarray}
The centered kernel vector is then mapped to
\begin{eqnarray}\label{npttest}
{\mathbf{\phi}}_{*} = \mathbf{(\Phi}^T)^{+}\mathbf{\Hat{k}}_{*}
\end{eqnarray}
We now consider ${\mathbf{\phi}}_{*}$ as the test input ${\mathbf{x}}_{*}$ and follow all the steps described for the linear test.
\section{Experiments}\label{experiments}
\subsection{Datasets and experimental setup}
We evaluated the proposed and competing methods over different datasets downloaded from UCI machine learning repository \cite{Dua:2019}. The datasets were converted to one-class datasets by considering a single class in the dataset at a time as the target class and all other classes as outliers. The total number of samples, number of target class samples, and number of dimensions in the original feature space are given in Table \ref{datasets}.
 \begin{table}[ht]
  \footnotesize\setlength{\tabcolsep}{10pt}
      \centering 
    \caption{Datasets used in the experiments} 
\begin{tabular}{|l|l|l|l|l|}
\hline
\textbf{Abbreviation} & \textbf{Dataset Name (Target Class)}    & \textbf{Total Samples} & \textbf{Target Samples} & \textbf{D} \\ \hline
S-K                & Seeds (Kama)                            & 210                    & 70                      & 7          \\ \hline
S-R                & Seeds (Rosa)                            & 210                    & 70                      & 7          \\ \hline
S-C                & Seeds (Canadian)                        & 210                    & 70                      & 7          \\ \hline
QB-B               & Qualitative bankruptcy (bankruptcy)     & 250                    & 107                     & 6          \\ \hline
QB-N               & Qualitative bankruptcy (non-bankruptcy) & 250                    & 143                     & 6          \\ \hline
SH-H               & Somerville happiness (happy)            & 143                    & 77                      & 6          \\ \hline
SH-U               & Somerville happiness (un-happy)         & 143                    & 66                      & 6          \\ \hline
I-S                & Iris (Setosa)                           & 150                    & 50                      & 4          \\ \hline
I-VC               & Iris (Versicolor)                       & 150                    & 50                      & 4          \\ \hline
I-V                & Iris (Virginica)                        & 150                    & 50                      & 4          \\ \hline
IS-B               & Ionosphere (bad)                        & 351                    & 126                     & 34         \\ \hline
IS-G               & Ionosphere (good)                       & 351                    & 225                     & 34         \\ \hline
SR-R               & Sonar (rock)                            & 208                    & 97                      & 60         \\ \hline
SR-M               & Sonar (mines)                           & 208                    & 111                     & 60         \\ \hline
\end{tabular}
\label{datasets}
\end{table}

 In each dataset, 70$\%$ of the data was used for training and the remaining 30$\%$ for testing. The train and test sets were selected randomly by keeping the proportions of classes similar to the full dataset. Each experiment was repeated five times using different random train/test splits, while the same five splittings were used for all the considered methods. We report the average test performance over the five splittings. During training, a 5-fold cross-validation technique was used to select the best hyperparameters with the best evaluation score. We used only the training sets for selecting the hyperparameters. We used Geometric mean (\textit{Gmean}) as the evaluation metric for all the methods. \textit{Gmean} is defined as
\begin{equation}\label{gmean}
 Gmean=\sqrt {tpr \times tnr},
\end{equation}
where $tpr$ is true positive rate (also known as sensitivity) and $tnr$ is true negative rate (also known as specificity).
 For the proposed ES-SVDD method, we chose the hyperparameters from the following values
\begin{itemize}
  \item$\beta\in\{10^{-4},10^{-3},10^{-2},10^{-1},10^{0},10^{1},10^{2},10^{3},10^{4}\},$
  \item$C\in\{0.01,0.05,0.1,0.2,0.3,0.4,0.5,0.6\}$,
  \item$\sigma\in\{10^{-3},10^{-2},10^{-1},10^{0},10^{1},10^{2},10^{3}\}$,
  \item$d\in\{1,2,3,4,5,10,20,50, 100\}$,
  \item$\eta\in\{10^{-5},10^{-4},10^{-3},10^{-2},10^{-1}\}$.
\end{itemize}
  
For all the competing methods, the hyperparameters corresponding to ES-SVDD hyperparameters were selected from the above values. For other hyperparameters, the same ranges were used as provided in the corresponding work. We used the full training set with the optimal hyperparameters for the final training and then tested with the test set.

We compared the proposed ES-SVDD with S-SVDD \cite{sohrab2018subspace}, OC-SVM \cite{scholkopfu1999sv}, SVDD \cite{tax2004support}, and E-SVDD. For non-linear methods, we employed NPT for ES-SVDD and S-SVDD, and kernel trick for other methods. For S-SVDD, different regularization terms were proposed and, hence, we compared with all the previously proposed variants of S-SVDD. We used LIBSVM \cite{CC01a} toolbox implementation for OC-SVM and SVDD. The proposed ES-SVDD along with S-SVDD and E-SVDD were implemented by the authors using Matlab by leveraging LIBSVM.

\subsection{Experimental results and discussion}

\begin{table}[ht]
\footnotesize\setlength{\tabcolsep}{5.2pt}
  \centering 
       \caption{\textit{Gmean} results for linear methods over different datasets}
\begin{tabular}{llllllllllllll}
\cline{1-1} \cline{3-6} \cline{8-10} \cline{12-14}
\multicolumn{1}{|l|}{Dataset} & \multicolumn{1}{l|}{} & \multicolumn{1}{l|}{S-K} & \multicolumn{1}{l|}{S-R} & \multicolumn{1}{l|}{S-C} & \multicolumn{1}{l|}{Av.} & \multicolumn{1}{l|}{} & \multicolumn{1}{l|}{QB-B}  & \multicolumn{1}{l|}{QB-N}  & \multicolumn{1}{l|}{Av.} & \multicolumn{1}{l|}{} & \multicolumn{1}{l|}{SH-H}   & \multicolumn{1}{l|}{SH-U}   & \multicolumn{1}{l|}{Av.} \\ \cline{1-1} \cline{3-6} \cline{8-10} \cline{12-14} 
\multicolumn{1}{|l|}{ES-SVDD $\psi_0 \Upsilon_0$}     & \multicolumn{1}{l|}{} & \multicolumn{1}{l|}{0.831}                          & \multicolumn{1}{l|}{0.910}                          & \multicolumn{1}{l|}{0.771}                          & \multicolumn{1}{l|}{0.837}                       & \multicolumn{1}{l|}{} & \multicolumn{1}{l|}{0.837}                         & \multicolumn{1}{l|}{0.261}                         & \multicolumn{1}{l|}{0.549}                       & \multicolumn{1}{l|}{} & \multicolumn{1}{l|}{0.412}                          & \multicolumn{1}{l|}{0.505}                          & \multicolumn{1}{l|}{0.459}                       \\ \cline{1-1} \cline{3-6} \cline{8-10} \cline{12-14} 
\multicolumn{1}{|l|}{ES-SVDD $\psi_1$}                & \multicolumn{1}{l|}{} & \multicolumn{1}{l|}{0.828}                          & \multicolumn{1}{l|}{0.892}                          & \multicolumn{1}{l|}{0.895}                          & \multicolumn{1}{l|}{0.872}                       & \multicolumn{1}{l|}{} & \multicolumn{1}{l|}{0.763}                         & \multicolumn{1}{l|}{0.457}                         & \multicolumn{1}{l|}{\textbf{0.610}}              & \multicolumn{1}{l|}{} & \multicolumn{1}{l|}{0.472}                          & \multicolumn{1}{l|}{0.470}                          & \multicolumn{1}{l|}{0.471}                       \\ \cline{1-1} \cline{3-6} \cline{8-10} \cline{12-14} 
\multicolumn{1}{|l|}{ES-SVDD $\psi_2$}                & \multicolumn{1}{l|}{} & \multicolumn{1}{l|}{0.820}                          & \multicolumn{1}{l|}{0.893}                          & \multicolumn{1}{l|}{0.873}                          & \multicolumn{1}{l|}{0.862}                       & \multicolumn{1}{l|}{} & \multicolumn{1}{l|}{0.849}                         & \multicolumn{1}{l|}{0.178}                         & \multicolumn{1}{l|}{0.514}                       & \multicolumn{1}{l|}{} & \multicolumn{1}{l|}{0.531}                          & \multicolumn{1}{l|}{0.513}                          & \multicolumn{1}{l|}{0.522}                       \\ \cline{1-1} \cline{3-6} \cline{8-10} \cline{12-14} 
\multicolumn{1}{|l|}{ES-SVDD $\psi_3$}                & \multicolumn{1}{l|}{} & \multicolumn{1}{l|}{0.793}                          & \multicolumn{1}{l|}{0.904}                          & \multicolumn{1}{l|}{0.871}                          & \multicolumn{1}{l|}{0.856}                       & \multicolumn{1}{l|}{} & \multicolumn{1}{l|}{0.898}                         & \multicolumn{1}{l|}{0.233}                         & \multicolumn{1}{l|}{0.565}                       & \multicolumn{1}{l|}{} & \multicolumn{1}{l|}{0.460}                          & \multicolumn{1}{l|}{0.345}                          & \multicolumn{1}{l|}{0.403}                       \\ \cline{1-1} \cline{3-6} \cline{8-10} \cline{12-14} 
\multicolumn{1}{|l|}{ES-SVDD $\Upsilon_1$}            & \multicolumn{1}{l|}{} & \multicolumn{1}{l|}{0.817}                          & \multicolumn{1}{l|}{0.920}                          & \multicolumn{1}{l|}{0.895}                          & \multicolumn{1}{l|}{0.878}                       & \multicolumn{1}{l|}{} & \multicolumn{1}{l|}{0.850}                         & \multicolumn{1}{l|}{0.318}                         & \multicolumn{1}{l|}{0.584}                       & \multicolumn{1}{l|}{} & \multicolumn{1}{l|}{0.463}                          & \multicolumn{1}{l|}{0.521}                          & \multicolumn{1}{l|}{0.492}                       \\ \cline{1-1} \cline{3-6} \cline{8-10} \cline{12-14} 
\multicolumn{1}{|l|}{ES-SVDD $\Upsilon_2$}            & \multicolumn{1}{l|}{} & \multicolumn{1}{l|}{0.842}                          & \multicolumn{1}{l|}{0.910}                          & \multicolumn{1}{l|}{\textbf{0.909}}                 & \multicolumn{1}{l|}{\textbf{0.887}}              & \multicolumn{1}{l|}{} & \multicolumn{1}{l|}{0.809}                         & \multicolumn{1}{l|}{0.305}                         & \multicolumn{1}{l|}{0.557}                       & \multicolumn{1}{l|}{} & \multicolumn{1}{l|}{\textbf{0.546}}                 & \multicolumn{1}{l|}{\textbf{0.533}}                 & \multicolumn{1}{l|}{\textbf{0.540}}              \\ \cline{1-1} \cline{3-6} \cline{8-10} \cline{12-14} 
\multicolumn{1}{|l|}{ES-SVDD $\Upsilon_3$}            & \multicolumn{1}{l|}{} & \multicolumn{1}{l|}{\textbf{0.848}}                 & \multicolumn{1}{l|}{0.881}                          & \multicolumn{1}{l|}{0.878}                          & \multicolumn{1}{l|}{0.869}                       & \multicolumn{1}{l|}{} & \multicolumn{1}{l|}{0.866}                         & \multicolumn{1}{l|}{0.334}                         & \multicolumn{1}{l|}{0.600}                       & \multicolumn{1}{l|}{} & \multicolumn{1}{l|}{0.488}                          & \multicolumn{1}{l|}{0.469}                          & \multicolumn{1}{l|}{0.479}                       \\ \cline{1-1} \cline{3-6} \cline{8-10} \cline{12-14} 
\multicolumn{1}{|l|}{S-SVDD $\psi_0$}                 & \multicolumn{1}{l|}{} & \multicolumn{1}{l|}{0.793}                          & \multicolumn{1}{l|}{0.864}                          & \multicolumn{1}{l|}{0.809}                          & \multicolumn{1}{l|}{0.822}                       & \multicolumn{1}{l|}{} & \multicolumn{1}{l|}{0.716}                         & \multicolumn{1}{l|}{0.505}                         & \multicolumn{1}{l|}{\textbf{0.610}}              & \multicolumn{1}{l|}{} & \multicolumn{1}{l|}{0.490}                          & \multicolumn{1}{l|}{0.478}                          & \multicolumn{1}{l|}{0.484}                       \\ \cline{1-1} \cline{3-6} \cline{8-10} \cline{12-14} 
\multicolumn{1}{|l|}{S-SVDD $\psi_1$}                 & \multicolumn{1}{l|}{} & \multicolumn{1}{l|}{0.717}                          & \multicolumn{1}{l|}{0.764}                          & \multicolumn{1}{l|}{0.769}                          & \multicolumn{1}{l|}{0.750}                       & \multicolumn{1}{l|}{} & \multicolumn{1}{l|}{0.852}                         & \multicolumn{1}{l|}{0.337}                         & \multicolumn{1}{l|}{0.595}                       & \multicolumn{1}{l|}{} & \multicolumn{1}{l|}{0.461}                          & \multicolumn{1}{l|}{0.460}                          & \multicolumn{1}{l|}{0.460}                       \\ \cline{1-1} \cline{3-6} \cline{8-10} \cline{12-14} 
\multicolumn{1}{|l|}{S-SVDD $\psi_2$}                 & \multicolumn{1}{l|}{} & \multicolumn{1}{l|}{0.809}                          & \multicolumn{1}{l|}{0.823}                          & \multicolumn{1}{l|}{0.773}                          & \multicolumn{1}{l|}{0.801}                       & \multicolumn{1}{l|}{} & \multicolumn{1}{l|}{0.755}                         & \multicolumn{1}{l|}{0.400}                         & \multicolumn{1}{l|}{0.577}                       & \multicolumn{1}{l|}{} & \multicolumn{1}{l|}{0.471}                          & \multicolumn{1}{l|}{0.481}                          & \multicolumn{1}{l|}{0.476}                       \\ \cline{1-1} \cline{3-6} \cline{8-10} \cline{12-14} 
\multicolumn{1}{|l|}{S-SVDD $\psi_3$}                 & \multicolumn{1}{l|}{} & \multicolumn{1}{l|}{0.804}                          & \multicolumn{1}{l|}{\textbf{0.930}}                 & \multicolumn{1}{l|}{0.751}                          & \multicolumn{1}{l|}{0.829}                       & \multicolumn{1}{l|}{} & \multicolumn{1}{l|}{0.724}                         & \multicolumn{1}{l|}{0.410}                         & \multicolumn{1}{l|}{0.567}                       & \multicolumn{1}{l|}{} & \multicolumn{1}{l|}{0.490}                          & \multicolumn{1}{l|}{0.464}                          & \multicolumn{1}{l|}{0.477}                       \\ \cline{1-1} \cline{3-6} \cline{8-10} \cline{12-14} 
\multicolumn{1}{|l|}{OC-SVM}                          & \multicolumn{1}{l|}{} & \multicolumn{1}{l|}{0.435}                          & \multicolumn{1}{l|}{0.458}                          & \multicolumn{1}{l|}{0.583}                          & \multicolumn{1}{l|}{0.492}                       & \multicolumn{1}{l|}{} & \multicolumn{1}{l|}{0.458}                         & \multicolumn{1}{l|}{\textbf{0.554}}                & \multicolumn{1}{l|}{0.506}                       & \multicolumn{1}{l|}{} & \multicolumn{1}{l|}{0.453}                          & \multicolumn{1}{l|}{0.424}                          & \multicolumn{1}{l|}{0.439}                       \\ \cline{1-1} \cline{3-6} \cline{8-10} \cline{12-14} 
\multicolumn{1}{|l|}{SVDD}                            & \multicolumn{1}{l|}{} & \multicolumn{1}{l|}{0.824}                          & \multicolumn{1}{l|}{0.917}                          & \multicolumn{1}{l|}{0.860}                          & \multicolumn{1}{l|}{0.867}                       & \multicolumn{1}{l|}{} & \multicolumn{1}{l|}{0.827}                         & \multicolumn{1}{l|}{0.043}                         & \multicolumn{1}{l|}{0.435}                       & \multicolumn{1}{l|}{} & \multicolumn{1}{l|}{0.536}                          & \multicolumn{1}{l|}{0.478}                          & \multicolumn{1}{l|}{0.507}                       \\ \cline{1-1} \cline{3-6} \cline{8-10} \cline{12-14} 
\multicolumn{1}{|l|}{E-SVDD}                         & \multicolumn{1}{l|}{} & \multicolumn{1}{l|}{0.802}                          & \multicolumn{1}{l|}{0.867}                          & \multicolumn{1}{l|}{0.864}                          & \multicolumn{1}{l|}{0.844}                       & \multicolumn{1}{l|}{} & \multicolumn{1}{l|}{\textbf{0.964}}                & \multicolumn{1}{l|}{0.202}                         & \multicolumn{1}{l|}{0.583}                       & \multicolumn{1}{l|}{} & \multicolumn{1}{l|}{0.543}                          & \multicolumn{1}{l|}{0.411}                          & \multicolumn{1}{l|}{0.477}                       \\ \cline{1-1} \cline{3-6} \cline{8-10} \cline{12-14} 
                                                      &                       &                                                     &                                                     &                                                     &                                                  &                       &                                                    &                                                    &                                                  &                       &                                                     &                                                     &                                                  \\ \cline{1-1} \cline{3-6} \cline{8-10} \cline{12-14} 
\multicolumn{1}{|l|}{Dataset} & \multicolumn{1}{l|}{} & \multicolumn{1}{l|}{I-S}  & \multicolumn{1}{l|}{I-VC}  & \multicolumn{1}{l|}{I-V}  & \multicolumn{1}{l|}{Av.} & \multicolumn{1}{l|}{} & \multicolumn{1}{l|}{IS-B} & \multicolumn{1}{l|}{IS-G} & \multicolumn{1}{l|}{Av.} & \multicolumn{1}{l|}{} & \multicolumn{1}{l|}{SR-R} & \multicolumn{1}{l|}{SR-M} & \multicolumn{1}{l|}{Av.} \\ \cline{1-1} \cline{3-6} \cline{8-10} \cline{12-14} 
\multicolumn{1}{|l|}{ES-SVDD $\psi_0 \Upsilon_0$}     & \multicolumn{1}{l|}{} & \multicolumn{1}{l|}{0.643}                          & \multicolumn{1}{l|}{0.749}                          & \multicolumn{1}{l|}{0.705}                          & \multicolumn{1}{l|}{0.699}                       & \multicolumn{1}{l|}{} & \multicolumn{1}{l|}{0.156}                         & \multicolumn{1}{l|}{0.889}                         & \multicolumn{1}{l|}{0.522}                       & \multicolumn{1}{l|}{} & \multicolumn{1}{l|}{\textbf{0.502}}                 & \multicolumn{1}{l|}{0.638}                          & \multicolumn{1}{l|}{0.570}                       \\ \cline{1-1} \cline{3-6} \cline{8-10} \cline{12-14} 
\multicolumn{1}{|l|}{ES-SVDD $\psi_1$}                & \multicolumn{1}{l|}{} & \multicolumn{1}{l|}{0.920}                          & \multicolumn{1}{l|}{0.858}                          & \multicolumn{1}{l|}{0.771}                          & \multicolumn{1}{l|}{0.850}                       & \multicolumn{1}{l|}{} & \multicolumn{1}{l|}{\textbf{0.522}}                & \multicolumn{1}{l|}{0.855}                         & \multicolumn{1}{l|}{\textbf{0.688}}              & \multicolumn{1}{l|}{} & \multicolumn{1}{l|}{0.501}                          & \multicolumn{1}{l|}{0.557}                          & \multicolumn{1}{l|}{0.529}                       \\ \cline{1-1} \cline{3-6} \cline{8-10} \cline{12-14} 
\multicolumn{1}{|l|}{ES-SVDD $\psi_2$}                & \multicolumn{1}{l|}{} & \multicolumn{1}{l|}{0.870}                          & \multicolumn{1}{l|}{0.821}                          & \multicolumn{1}{l|}{0.793}                          & \multicolumn{1}{l|}{0.828}                       & \multicolumn{1}{l|}{} & \multicolumn{1}{l|}{0.309}                         & \multicolumn{1}{l|}{0.875}                         & \multicolumn{1}{l|}{0.592}                       & \multicolumn{1}{l|}{} & \multicolumn{1}{l|}{0.480}                          & \multicolumn{1}{l|}{\textbf{0.671}}                 & \multicolumn{1}{l|}{\textbf{0.575}}              \\ \cline{1-1} \cline{3-6} \cline{8-10} \cline{12-14} 
\multicolumn{1}{|l|}{ES-SVDD $\psi_3$}                & \multicolumn{1}{l|}{} & \multicolumn{1}{l|}{0.927}                          & \multicolumn{1}{l|}{0.874}                          & \multicolumn{1}{l|}{0.710}                          & \multicolumn{1}{l|}{0.837}                       & \multicolumn{1}{l|}{} & \multicolumn{1}{l|}{0.353}                         & \multicolumn{1}{l|}{0.887}                         & \multicolumn{1}{l|}{0.620}                       & \multicolumn{1}{l|}{} & \multicolumn{1}{l|}{0.480}                          & \multicolumn{1}{l|}{0.654}                          & \multicolumn{1}{l|}{0.567}                       \\ \cline{1-1} \cline{3-6} \cline{8-10} \cline{12-14} 
\multicolumn{1}{|l|}{ES-SVDD $\Upsilon_1$}            & \multicolumn{1}{l|}{} & \multicolumn{1}{l|}{0.851}                          & \multicolumn{1}{l|}{0.842}                          & \multicolumn{1}{l|}{0.861}                          & \multicolumn{1}{l|}{0.851}                       & \multicolumn{1}{l|}{} & \multicolumn{1}{l|}{0.263}                         & \multicolumn{1}{l|}{0.871}                         & \multicolumn{1}{l|}{0.567}                       & \multicolumn{1}{l|}{} & \multicolumn{1}{l|}{0.472}                          & \multicolumn{1}{l|}{0.668}                          & \multicolumn{1}{l|}{0.570}                       \\ \cline{1-1} \cline{3-6} \cline{8-10} \cline{12-14} 
\multicolumn{1}{|l|}{ES-SVDD $\Upsilon_2$}            & \multicolumn{1}{l|}{} & \multicolumn{1}{l|}{\textbf{0.958}}                 & \multicolumn{1}{l|}{0.828}                          & \multicolumn{1}{l|}{0.742}                          & \multicolumn{1}{l|}{0.843}                       & \multicolumn{1}{l|}{} & \multicolumn{1}{l|}{0.311}                         & \multicolumn{1}{l|}{0.886}                         & \multicolumn{1}{l|}{0.598}                       & \multicolumn{1}{l|}{} & \multicolumn{1}{l|}{0.468}                          & \multicolumn{1}{l|}{0.651}                          & \multicolumn{1}{l|}{0.560}                       \\ \cline{1-1} \cline{3-6} \cline{8-10} \cline{12-14} 
\multicolumn{1}{|l|}{ES-SVDD $\Upsilon_3$}            & \multicolumn{1}{l|}{} & \multicolumn{1}{l|}{0.802}                          & \multicolumn{1}{l|}{0.849}                          & \multicolumn{1}{l|}{0.790}                          & \multicolumn{1}{l|}{0.814}                       & \multicolumn{1}{l|}{} & \multicolumn{1}{l|}{0.345}                         & \multicolumn{1}{l|}{\textbf{0.901}}                & \multicolumn{1}{l|}{0.623}                       & \multicolumn{1}{l|}{} & \multicolumn{1}{l|}{0.495}                          & \multicolumn{1}{l|}{0.651}                          & \multicolumn{1}{l|}{0.573}                       \\ \cline{1-1} \cline{3-6} \cline{8-10} \cline{12-14} 
\multicolumn{1}{|l|}{S-SVDD $\psi_0$}                 & \multicolumn{1}{l|}{} & \multicolumn{1}{l|}{0.870}                          & \multicolumn{1}{l|}{0.752}                          & \multicolumn{1}{l|}{0.644}                          & \multicolumn{1}{l|}{0.756}                       & \multicolumn{1}{l|}{} & \multicolumn{1}{l|}{0.164}                         & \multicolumn{1}{l|}{0.749}                         & \multicolumn{1}{l|}{0.456}                       & \multicolumn{1}{l|}{} & \multicolumn{1}{l|}{0.373}                          & \multicolumn{1}{l|}{0.374}                          & \multicolumn{1}{l|}{0.373}                       \\ \cline{1-1} \cline{3-6} \cline{8-10} \cline{12-14} 
\multicolumn{1}{|l|}{S-SVDD $\psi_1$}                 & \multicolumn{1}{l|}{} & \multicolumn{1}{l|}{0.882}                          & \multicolumn{1}{l|}{0.809}                          & \multicolumn{1}{l|}{0.752}                          & \multicolumn{1}{l|}{0.814}                       & \multicolumn{1}{l|}{} & \multicolumn{1}{l|}{0.504}                         & \multicolumn{1}{l|}{0.706}                         & \multicolumn{1}{l|}{0.605}                       & \multicolumn{1}{l|}{} & \multicolumn{1}{l|}{0.444}                          & \multicolumn{1}{l|}{0.358}                          & \multicolumn{1}{l|}{0.401}                       \\ \cline{1-1} \cline{3-6} \cline{8-10} \cline{12-14} 
\multicolumn{1}{|l|}{S-SVDD $\psi_2$}                 & \multicolumn{1}{l|}{} & \multicolumn{1}{l|}{0.874}                          & \multicolumn{1}{l|}{0.844}                          & \multicolumn{1}{l|}{0.576}                          & \multicolumn{1}{l|}{0.764}                       & \multicolumn{1}{l|}{} & \multicolumn{1}{l|}{0.433}                         & \multicolumn{1}{l|}{0.719}                         & \multicolumn{1}{l|}{0.576}                       & \multicolumn{1}{l|}{} & \multicolumn{1}{l|}{0.463}                          & \multicolumn{1}{l|}{0.398}                          & \multicolumn{1}{l|}{0.430}                       \\ \cline{1-1} \cline{3-6} \cline{8-10} \cline{12-14} 
\multicolumn{1}{|l|}{S-SVDD $\psi_3$}                 & \multicolumn{1}{l|}{} & \multicolumn{1}{l|}{0.812}                          & \multicolumn{1}{l|}{0.680}                          & \multicolumn{1}{l|}{0.633}                          & \multicolumn{1}{l|}{0.708}                       & \multicolumn{1}{l|}{} & \multicolumn{1}{l|}{0.271}                         & \multicolumn{1}{l|}{0.657}                         & \multicolumn{1}{l|}{0.464}                       & \multicolumn{1}{l|}{} & \multicolumn{1}{l|}{0.459}                          & \multicolumn{1}{l|}{0.409}                          & \multicolumn{1}{l|}{0.434}                       \\ \cline{1-1} \cline{3-6} \cline{8-10} \cline{12-14} 
\multicolumn{1}{|l|}{OC-SVM}                          & \multicolumn{1}{l|}{} & \multicolumn{1}{l|}{0.496}                          & \multicolumn{1}{l|}{0.521}                          & \multicolumn{1}{l|}{0.390}                          & \multicolumn{1}{l|}{0.469}                       & \multicolumn{1}{l|}{} & \multicolumn{1}{l|}{0.473}                         & \multicolumn{1}{l|}{0.453}                         & \multicolumn{1}{l|}{0.463}                       & \multicolumn{1}{l|}{} & \multicolumn{1}{l|}{0.443}                          & \multicolumn{1}{l|}{0.517}                          & \multicolumn{1}{l|}{0.480}                       \\ \cline{1-1} \cline{3-6} \cline{8-10} \cline{12-14} 
\multicolumn{1}{|l|}{SVDD}                            & \multicolumn{1}{l|}{} & \multicolumn{1}{l|}{0.945}                          & \multicolumn{1}{l|}{\textbf{0.896}}                 & \multicolumn{1}{l|}{\textbf{0.893}}                 & \multicolumn{1}{l|}{\textbf{0.911}}              & \multicolumn{1}{l|}{} & \multicolumn{1}{l|}{0.043}                         & \multicolumn{1}{l|}{0.727}                         & \multicolumn{1}{l|}{0.385}                       & \multicolumn{1}{l|}{} & \multicolumn{1}{l|}{0.498}                          & \multicolumn{1}{l|}{0.520}                          & \multicolumn{1}{l|}{0.509}                       \\ \cline{1-1} \cline{3-6} \cline{8-10} \cline{12-14} 
\multicolumn{1}{|l|}{E-SVDD}                         & \multicolumn{1}{l|}{} & \multicolumn{1}{l|}{0.887}                          & \multicolumn{1}{l|}{0.879}                          & \multicolumn{1}{l|}{0.862}                          & \multicolumn{1}{l|}{0.876}                       & \multicolumn{1}{l|}{} & \multicolumn{1}{l|}{0.335}                         & \multicolumn{1}{l|}{0.000}                         & \multicolumn{1}{l|}{0.167}                       & \multicolumn{1}{l|}{} & \multicolumn{1}{l|}{0.000}                          & \multicolumn{1}{l|}{0.000}                          & \multicolumn{1}{l|}{0.000}                       \\ \cline{1-1} \cline{3-6} \cline{8-10} \cline{12-14} 
\end{tabular}\label{linearresults}  
\end{table}

In Tables \ref{linearresults} and \ref{kernelresults}, we report the average test results for each dataset for the linear and non-linear cases, respectively. In each experiment, a single class was selected as the target class and the rest of the data as outliers (see Table \ref{datasets}). We also report the average performance of the proposed and competing methods in the average (Av.) column by averaging the results for a given dataset. For example, the performance over S-K, S-R, and S-C is averaged and provided in the Av. column as the overall performance for Seeds dataset. In this way, we can get an idea of the overall performance for each algorithm over the full dataset. For ES-SVDD and S-SVDD, we report the test results after 10 training iterations. 

In the linear case, our proposed methods achieved the best average results on all but Iris dataset and, in the non-linear, the proposed methods achieved the best averages on half of the datasets. We note that the average results for the non-linear methods are generally better than those of the linear ones in the majority of the datasets. Overall, the proposed (linear and non-linear) methods achieve the best average results in 4 out of 6 datasets. In general, the best performing methods vary for different datasets, but we can see that there is no case, where the proposed method would fail completely, unlike most of the competing methods. In the linear case, other competing methods outperformed ES-SVDD only with Iris dataset, which has the lowest original dimensionality and also a low number of samples. Also in the non-linear case, other methods outperformed ES-SVDD most clearly on the 2 smallest datasets. Thus, it seems that the proposed method is more beneficial when the data dimensionality is higher.

Comparing regularization terms for linear ES-SVDD, we notice that ES-SVDD performs better in majority of cases with regularization term $\Upsilon_2$ which uses samples belonging to the boundary and outside the boundary to describe the covariance of the class. Regularization term $\psi_1$ which uses all training samples to describe the covariance of the class, also performs well. Both of these regularization terms produced 2 out of 6 best results in the linear case. We also noticed that ES-SVDD without any regularization term performs the worst as compared to ES-SVDD with regularization terms.  

For non-linear ES-SVDD, the regularization terms $\psi_1$ and $\Upsilon_3$ resulted in the best results for most of the datasets. However, $\psi_1$ is also noticed to perform worse than the others in a few datasets. $\psi_1$ uses all target training samples in describing the covariance of the class without taking into account the concentration matrix. In $\Upsilon_3$, the $\boldsymbol{\lambda}$ values take the values of $\alpha_i$ values of the support vectors and zero for non-support vectors. In the non-linear case for high dimensional datasets, we notice that using all the training data for describing the covariance of the data in a projected space, with or without using the concentration matrix (i.e., $\psi_1$ or $\Upsilon_1$), yielded the best results for ES-SVDD.

\begin{table}[ht]
\footnotesize\setlength{\tabcolsep}{5.2pt}
  \centering 
       \caption{\textit{Gmean} results for \textbf{non-linear} methods over different datasets}
\begin{tabular}{llllllllllllll}
\cline{1-1} \cline{3-6} \cline{8-10} \cline{12-14}
\multicolumn{1}{|l|}{Dataset} & \multicolumn{1}{l|}{} & \multicolumn{1}{l|}{S-K} & \multicolumn{1}{l|}{S-R} & \multicolumn{1}{l|}{S-C} & \multicolumn{1}{l|}{Av.} & \multicolumn{1}{l|}{} & \multicolumn{1}{l|}{QB-B}  & \multicolumn{1}{l|}{QB-N}  & \multicolumn{1}{l|}{Av.} & \multicolumn{1}{l|}{} & \multicolumn{1}{l|}{SH-H}   & \multicolumn{1}{l|}{SH-U}   & \multicolumn{1}{l|}{Av.} \\ \cline{1-1} \cline{3-6} \cline{8-10} \cline{12-14} 
\multicolumn{1}{|l|}{ES-SVDD $\psi_0 \Upsilon_0$}     & \multicolumn{1}{l|}{} & \multicolumn{1}{l|}{0.783}                          & \multicolumn{1}{l|}{0.884}                          & \multicolumn{1}{l|}{0.\textbf{928}}                          & \multicolumn{1}{l|}{0.865}                       & \multicolumn{1}{l|}{} & \multicolumn{1}{l|}{0.829}                         & \multicolumn{1}{l|}{0.613}                         & \multicolumn{1}{l|}{0.721}                       & \multicolumn{1}{l|}{} & \multicolumn{1}{l|}{0.516}                          & \multicolumn{1}{l|}{0.423}                          & \multicolumn{1}{l|}{0.469}                       \\ \cline{1-1} \cline{3-6} \cline{8-10} \cline{12-14} 
\multicolumn{1}{|l|}{ES-SVDD $\psi_1$}                & \multicolumn{1}{l|}{} & \multicolumn{1}{l|}{0.802}                          & \multicolumn{1}{l|}{0.878}                          & \multicolumn{1}{l|}{0.882}                          & \multicolumn{1}{l|}{0.854}                       & \multicolumn{1}{l|}{} & \multicolumn{1}{l|}{0.804}                         & \multicolumn{1}{l|}{0.343}                         & \multicolumn{1}{l|}{0.573}                       & \multicolumn{1}{l|}{} & \multicolumn{1}{l|}{0.512}                          & \multicolumn{1}{l|}{0.423}                          & \multicolumn{1}{l|}{0.467}                       \\ \cline{1-1} \cline{3-6} \cline{8-10} \cline{12-14} 
\multicolumn{1}{|l|}{ES-SVDD $\psi_2$}                & \multicolumn{1}{l|}{} & \multicolumn{1}{l|}{0.797}                          & \multicolumn{1}{l|}{0.898}                          & \multicolumn{1}{l|}{\textbf{0.928}}                 & \multicolumn{1}{l|}{0.875}                       & \multicolumn{1}{l|}{} & \multicolumn{1}{l|}{0.895}                         & \multicolumn{1}{l|}{0.345}                         & \multicolumn{1}{l|}{0.620}                       & \multicolumn{1}{l|}{} & \multicolumn{1}{l|}{0.517}                          & \multicolumn{1}{l|}{0.330}                          & \multicolumn{1}{l|}{0.423}                       \\ \cline{1-1} \cline{3-6} \cline{8-10} \cline{12-14} 
\multicolumn{1}{|l|}{ES-SVDD $\psi_3$}                & \multicolumn{1}{l|}{} & \multicolumn{1}{l|}{0.816}                          & \multicolumn{1}{l|}{0.860}                          & \multicolumn{1}{l|}{0.721}                          & \multicolumn{1}{l|}{0.799}                       & \multicolumn{1}{l|}{} & \multicolumn{1}{l|}{0.892}                         & \multicolumn{1}{l|}{0.636}                         & \multicolumn{1}{l|}{\textbf{0.764}}              & \multicolumn{1}{l|}{} & \multicolumn{1}{l|}{0.520}                          & \multicolumn{1}{l|}{0.450}                          & \multicolumn{1}{l|}{0.485}                       \\ \cline{1-1} \cline{3-6} \cline{8-10} \cline{12-14} 
\multicolumn{1}{|l|}{ES-SVDD $\Upsilon_1$}            & \multicolumn{1}{l|}{} & \multicolumn{1}{l|}{0.847}                          & \multicolumn{1}{l|}{\textbf{0.916}}                 & \multicolumn{1}{l|}{0.911}                          & \multicolumn{1}{l|}{\textbf{0.891}}              & \multicolumn{1}{l|}{} & \multicolumn{1}{l|}{0.871}                         & \multicolumn{1}{l|}{0.467}                         & \multicolumn{1}{l|}{0.669}                       & \multicolumn{1}{l|}{} & \multicolumn{1}{l|}{0.474}                          & \multicolumn{1}{l|}{0.380}                          & \multicolumn{1}{l|}{0.427}                       \\ \cline{1-1} \cline{3-6} \cline{8-10} \cline{12-14} 
\multicolumn{1}{|l|}{ES-SVDD $\Upsilon_2$}            & \multicolumn{1}{l|}{} & \multicolumn{1}{l|}{0.817}                          & \multicolumn{1}{l|}{0.879}                          & \multicolumn{1}{l|}{0.889}                          & \multicolumn{1}{l|}{0.862}                       & \multicolumn{1}{l|}{} & \multicolumn{1}{l|}{0.842}                         & \multicolumn{1}{l|}{\textbf{0.679}}                & \multicolumn{1}{l|}{0.760}                       & \multicolumn{1}{l|}{} & \multicolumn{1}{l|}{0.516}                          & \multicolumn{1}{l|}{0.342}                          & \multicolumn{1}{l|}{0.429}                       \\ \cline{1-1} \cline{3-6} \cline{8-10} \cline{12-14} 
\multicolumn{1}{|l|}{ES-SVDD $\Upsilon_3$}            & \multicolumn{1}{l|}{} & \multicolumn{1}{l|}{0.847}                          & \multicolumn{1}{l|}{0.885}                          & \multicolumn{1}{l|}{0.908}                          & \multicolumn{1}{l|}{0.880}                       & \multicolumn{1}{l|}{} & \multicolumn{1}{l|}{0.871}                         & \multicolumn{1}{l|}{0.541}                         & \multicolumn{1}{l|}{0.706}                       & \multicolumn{1}{l|}{} & \multicolumn{1}{l|}{0.523}                          & \multicolumn{1}{l|}{0.449}                          & \multicolumn{1}{l|}{0.486}                       \\ \cline{1-1} \cline{3-6} \cline{8-10} \cline{12-14} 
\multicolumn{1}{|l|}{S-SVDD $\psi_0$}                 & \multicolumn{1}{l|}{} & \multicolumn{1}{l|}{0.737}                          & \multicolumn{1}{l|}{0.742}                          & \multicolumn{1}{l|}{0.831}                          & \multicolumn{1}{l|}{0.770}                       & \multicolumn{1}{l|}{} & \multicolumn{1}{l|}{0.226}                         & \multicolumn{1}{l|}{0.488}                         & \multicolumn{1}{l|}{0.357}                       & \multicolumn{1}{l|}{} & \multicolumn{1}{l|}{0.452}                          & \multicolumn{1}{l|}{0.289}                          & \multicolumn{1}{l|}{0.371}                       \\ \cline{1-1} \cline{3-6} \cline{8-10} \cline{12-14} 
\multicolumn{1}{|l|}{S-SVDD $\psi_1$}                 & \multicolumn{1}{l|}{} & \multicolumn{1}{l|}{0.714}                          & \multicolumn{1}{l|}{0.775}                          & \multicolumn{1}{l|}{0.815}                          & \multicolumn{1}{l|}{0.768}                       & \multicolumn{1}{l|}{} & \multicolumn{1}{l|}{0.110}                         & \multicolumn{1}{l|}{0.077}                         & \multicolumn{1}{l|}{0.094}                       & \multicolumn{1}{l|}{} & \multicolumn{1}{l|}{0.386}                          & \multicolumn{1}{l|}{0.317}                          & \multicolumn{1}{l|}{0.352}                       \\ \cline{1-1} \cline{3-6} \cline{8-10} \cline{12-14} 
\multicolumn{1}{|l|}{S-SVDD $\psi_2$}                 & \multicolumn{1}{l|}{} & \multicolumn{1}{l|}{0.724}                          & \multicolumn{1}{l|}{0.854}                          & \multicolumn{1}{l|}{0.829}                          & \multicolumn{1}{l|}{0.802}                       & \multicolumn{1}{l|}{} & \multicolumn{1}{l|}{0.360}                         & \multicolumn{1}{l|}{0.370}                         & \multicolumn{1}{l|}{0.365}                       & \multicolumn{1}{l|}{} & \multicolumn{1}{l|}{0.474}                          & \multicolumn{1}{l|}{0.320}                          & \multicolumn{1}{l|}{0.397}                       \\ \cline{1-1} \cline{3-6} \cline{8-10} \cline{12-14} 
\multicolumn{1}{|l|}{S-SVDD $\psi_3$}                 & \multicolumn{1}{l|}{} & \multicolumn{1}{l|}{0.601}                          & \multicolumn{1}{l|}{0.757}                          & \multicolumn{1}{l|}{0.763}                          & \multicolumn{1}{l|}{0.707}                       & \multicolumn{1}{l|}{} & \multicolumn{1}{l|}{0.360}                         & \multicolumn{1}{l|}{0.401}                         & \multicolumn{1}{l|}{0.381}                       & \multicolumn{1}{l|}{} & \multicolumn{1}{l|}{0.460}                          & \multicolumn{1}{l|}{0.289}                          & \multicolumn{1}{l|}{0.375}                       \\ \cline{1-1} \cline{3-6} \cline{8-10} \cline{12-14} 
\multicolumn{1}{|l|}{OC-SVM}                          & \multicolumn{1}{l|}{} & \multicolumn{1}{l|}{0.793}                          & \multicolumn{1}{l|}{0.597}                          & \multicolumn{1}{l|}{0.631}                          & \multicolumn{1}{l|}{0.674}                       & \multicolumn{1}{l|}{} & \multicolumn{1}{l|}{0.671}                         & \multicolumn{1}{l|}{0.518}                         & \multicolumn{1}{l|}{0.594}                       & \multicolumn{1}{l|}{} & \multicolumn{1}{l|}{\textbf{0.571}}                 & \multicolumn{1}{l|}{0.478}                          & \multicolumn{1}{l|}{\textbf{0.525}}              \\ \cline{1-1} \cline{3-6} \cline{8-10} \cline{12-14} 
\multicolumn{1}{|l|}{SVDD}                            & \multicolumn{1}{l|}{} & \multicolumn{1}{l|}{\textbf{0.858}}                 & \multicolumn{1}{l|}{0.913}                          & \multicolumn{1}{l|}{0.879}                          & \multicolumn{1}{l|}{0.883}                       & \multicolumn{1}{l|}{} & \multicolumn{1}{l|}{0.876}                         & \multicolumn{1}{l|}{0.551}                         & \multicolumn{1}{l|}{0.714}                       & \multicolumn{1}{l|}{} & \multicolumn{1}{l|}{0.542}                          & \multicolumn{1}{l|}{\textbf{0.479}}                          & \multicolumn{1}{l|}{0.510}                       \\ \cline{1-1} \cline{3-6} \cline{8-10} \cline{12-14} 
\multicolumn{1}{|l|}{E-SVDD}                         & \multicolumn{1}{l|}{} & \multicolumn{1}{l|}{0.841}                          & \multicolumn{1}{l|}{0.853}                          & \multicolumn{1}{l|}{0.847}                          & \multicolumn{1}{l|}{0.847}                       & \multicolumn{1}{l|}{} & \multicolumn{1}{l|}{\textbf{0.961}}                & \multicolumn{1}{l|}{0.510}                         & \multicolumn{1}{l|}{0.736}                       & \multicolumn{1}{l|}{} & \multicolumn{1}{l|}{0.546}                          & \multicolumn{1}{l|}{0.418}                          & \multicolumn{1}{l|}{0.482}                       \\ \cline{1-1} \cline{3-6} \cline{8-10} \cline{12-14} 
\multicolumn{1}{|l|}{GE-SVDD}                         & \multicolumn{1}{l|}{} & \multicolumn{1}{l|}{0.835}                          & \multicolumn{1}{l|}{0.898}                          & \multicolumn{1}{l|}{0.757}                          & \multicolumn{1}{l|}{0.830}                       & \multicolumn{1}{l|}{} & \multicolumn{1}{l|}{0.936}                         & \multicolumn{1}{l|}{0.166}                         & \multicolumn{1}{l|}{0.551}                       & \multicolumn{1}{l|}{} & \multicolumn{1}{l|}{0.537}                          & \multicolumn{1}{l|}{{0.468}}                 & \multicolumn{1}{l|}{0.502}                       \\ \cline{1-1} \cline{3-6} \cline{8-10} \cline{12-14} 
\multicolumn{1}{|l|}{GE-SVM}                          & \multicolumn{1}{l|}{} & \multicolumn{1}{l|}{0.830}                          & \multicolumn{1}{l|}{0.880}                          & \multicolumn{1}{l|}{0.893}                          & \multicolumn{1}{l|}{0.867}                       & \multicolumn{1}{l|}{} & \multicolumn{1}{l|}{0.879}                         & \multicolumn{1}{l|}{0.577}                         & \multicolumn{1}{l|}{0.728}                       & \multicolumn{1}{l|}{} & \multicolumn{1}{l|}{\textbf{0.571}}                          & \multicolumn{1}{l|}{0.422}                          & \multicolumn{1}{l|}{0.496}                       \\ \cline{1-1} \cline{3-6} \cline{8-10} \cline{12-14} 
                                                      &                       &                                                     &                                                     &                                                     &                                                  &                       &                                                    &                                                    &                                                  &                       &                                                     &                                                     &                                                  \\ \cline{1-1} \cline{3-6} \cline{8-10} \cline{12-14} 
\multicolumn{1}{|l|}{Dataset} & \multicolumn{1}{l|}{} & \multicolumn{1}{l|}{I-S}  & \multicolumn{1}{l|}{I-VC}  & \multicolumn{1}{l|}{I-V}  & \multicolumn{1}{l|}{Av.} & \multicolumn{1}{l|}{} & \multicolumn{1}{l|}{IS-B} & \multicolumn{1}{l|}{IS-G} & \multicolumn{1}{l|}{Av.} & \multicolumn{1}{l|}{} & \multicolumn{1}{l|}{SR-R} & \multicolumn{1}{l|}{SR-M} & \multicolumn{1}{l|}{Av.} \\ \cline{1-1} \cline{3-6} \cline{8-10} \cline{12-14} 
\multicolumn{1}{|l|}{ES-SVDD $\psi_0 \Upsilon_0$}     & \multicolumn{1}{l|}{} & \multicolumn{1}{l|}{0.929}                          & \multicolumn{1}{l|}{0.843}                          & \multicolumn{1}{l|}{\textbf{0.864}}                 & \multicolumn{1}{l|}{0.879}                       & \multicolumn{1}{l|}{} & \multicolumn{1}{l|}{0.441}                         & \multicolumn{1}{l|}{0.894}                         & \multicolumn{1}{l|}{0.667}                       & \multicolumn{1}{l|}{} & \multicolumn{1}{l|}{0.410}                          & \multicolumn{1}{l|}{0.674}                          & \multicolumn{1}{l|}{0.542}                       \\ \cline{1-1} \cline{3-6} \cline{8-10} \cline{12-14} 
\multicolumn{1}{|l|}{ES-SVDD $\psi_1$}                & \multicolumn{1}{l|}{} & \multicolumn{1}{l|}{\textbf{0.938}}                 & \multicolumn{1}{l|}{0.811}                          & \multicolumn{1}{l|}{0.736}                          & \multicolumn{1}{l|}{0.829}                       & \multicolumn{1}{l|}{} & \multicolumn{1}{l|}{\textbf{0.710}}                & \multicolumn{1}{l|}{0.898}                         & \multicolumn{1}{l|}{\textbf{0.804}}              & \multicolumn{1}{l|}{} & \multicolumn{1}{l|}{0.480}                          & \multicolumn{1}{l|}{0.547}                          & \multicolumn{1}{l|}{0.514}                       \\ \cline{1-1} \cline{3-6} \cline{8-10} \cline{12-14} 
\multicolumn{1}{|l|}{ES-SVDD $\psi_2$}                & \multicolumn{1}{l|}{} & \multicolumn{1}{l|}{0.907}                          & \multicolumn{1}{l|}{0.868}                          & \multicolumn{1}{l|}{0.830}                          & \multicolumn{1}{l|}{0.868}                       & \multicolumn{1}{l|}{} & \multicolumn{1}{l|}{0.307}                         & \multicolumn{1}{l|}{0.865}                         & \multicolumn{1}{l|}{0.586}                       & \multicolumn{1}{l|}{} & \multicolumn{1}{l|}{0.471}                          & \multicolumn{1}{l|}{0.657}                          & \multicolumn{1}{l|}{0.564}                       \\ \cline{1-1} \cline{3-6} \cline{8-10} \cline{12-14} 
\multicolumn{1}{|l|}{ES-SVDD $\psi_3$}                & \multicolumn{1}{l|}{} & \multicolumn{1}{l|}{0.888}                          & \multicolumn{1}{l|}{0.843}                          & \multicolumn{1}{l|}{0.737}                          & \multicolumn{1}{l|}{0.823}                       & \multicolumn{1}{l|}{} & \multicolumn{1}{l|}{0.325}                         & \multicolumn{1}{l|}{0.883}                         & \multicolumn{1}{l|}{0.604}                       & \multicolumn{1}{l|}{} & \multicolumn{1}{l|}{0.469}                          & \multicolumn{1}{l|}{0.664}                          & \multicolumn{1}{l|}{0.567}                       \\ \cline{1-1} \cline{3-6} \cline{8-10} \cline{12-14} 
\multicolumn{1}{|l|}{ES-SVDD $\Upsilon_1$}            & \multicolumn{1}{l|}{} & \multicolumn{1}{l|}{0.814}                          & \multicolumn{1}{l|}{0.888}                          & \multicolumn{1}{l|}{0.702}                          & \multicolumn{1}{l|}{0.801}                       & \multicolumn{1}{l|}{} & \multicolumn{1}{l|}{0.469}                         & \multicolumn{1}{l|}{0.863}                         & \multicolumn{1}{l|}{0.666}                       & \multicolumn{1}{l|}{} & \multicolumn{1}{l|}{0.532}                          & \multicolumn{1}{l|}{0.648}                          & \multicolumn{1}{l|}{0.590}                       \\ \cline{1-1} \cline{3-6} \cline{8-10} \cline{12-14} 
\multicolumn{1}{|l|}{ES-SVDD $\Upsilon_2$}            & \multicolumn{1}{l|}{} & \multicolumn{1}{l|}{0.908}                          & \multicolumn{1}{l|}{0.833}                          & \multicolumn{1}{l|}{0.814}                          & \multicolumn{1}{l|}{0.851}                       & \multicolumn{1}{l|}{} & \multicolumn{1}{l|}{0.683}                         & \multicolumn{1}{l|}{0.855}                         & \multicolumn{1}{l|}{0.769}                       & \multicolumn{1}{l|}{} & \multicolumn{1}{l|}{0.470}                          & \multicolumn{1}{l|}{0.697}                          & \multicolumn{1}{l|}{0.583}                       \\ \cline{1-1} \cline{3-6} \cline{8-10} \cline{12-14} 
\multicolumn{1}{|l|}{ES-SVDD $\Upsilon_3$}            & \multicolumn{1}{l|}{} & \multicolumn{1}{l|}{\textbf{0.938}}                 & \multicolumn{1}{l|}{0.878}                          & \multicolumn{1}{l|}{0.828}                          & \multicolumn{1}{l|}{0.881}                       & \multicolumn{1}{l|}{} & \multicolumn{1}{l|}{0.452}                         & \multicolumn{1}{l|}{0.853}                         & \multicolumn{1}{l|}{0.653}                       & \multicolumn{1}{l|}{} & \multicolumn{1}{l|}{0.405}                          & \multicolumn{1}{l|}{\textbf{0.702}}                 & \multicolumn{1}{l|}{0.554}                       \\ \cline{1-1} \cline{3-6} \cline{8-10} \cline{12-14} 
\multicolumn{1}{|l|}{S-SVDD $\psi_0$}                 & \multicolumn{1}{l|}{} & \multicolumn{1}{l|}{0.916}                          & \multicolumn{1}{l|}{0.848}                          & \multicolumn{1}{l|}{0.776}                          & \multicolumn{1}{l|}{0.847}                       & \multicolumn{1}{l|}{} & \multicolumn{1}{l|}{0.238}                         & \multicolumn{1}{l|}{0.531}                         & \multicolumn{1}{l|}{0.384}                       & \multicolumn{1}{l|}{} & \multicolumn{1}{l|}{0.428}                          & \multicolumn{1}{l|}{0.408}                          & \multicolumn{1}{l|}{0.418}                       \\ \cline{1-1} \cline{3-6} \cline{8-10} \cline{12-14} 
\multicolumn{1}{|l|}{S-SVDD $\psi_1$}                 & \multicolumn{1}{l|}{} & \multicolumn{1}{l|}{0.893}                          & \multicolumn{1}{l|}{0.888}                          & \multicolumn{1}{l|}{0.627}                          & \multicolumn{1}{l|}{0.803}                       & \multicolumn{1}{l|}{} & \multicolumn{1}{l|}{0.684}                         & \multicolumn{1}{l|}{0.641}                         & \multicolumn{1}{l|}{0.662}                       & \multicolumn{1}{l|}{} & \multicolumn{1}{l|}{0.201}                          & \multicolumn{1}{l|}{0.485}                          & \multicolumn{1}{l|}{0.343}                       \\ \cline{1-1} \cline{3-6} \cline{8-10} \cline{12-14} 
\multicolumn{1}{|l|}{S-SVDD $\psi_2$}                 & \multicolumn{1}{l|}{} & \multicolumn{1}{l|}{0.914}                          & \multicolumn{1}{l|}{0.842}                          & \multicolumn{1}{l|}{0.773}                          & \multicolumn{1}{l|}{0.843}                       & \multicolumn{1}{l|}{} & \multicolumn{1}{l|}{0.208}                         & \multicolumn{1}{l|}{0.606}                         & \multicolumn{1}{l|}{0.407}                       & \multicolumn{1}{l|}{} & \multicolumn{1}{l|}{0.402}                          & \multicolumn{1}{l|}{0.522}                          & \multicolumn{1}{l|}{0.462}                       \\ \cline{1-1} \cline{3-6} \cline{8-10} \cline{12-14} 
\multicolumn{1}{|l|}{S-SVDD $\psi_3$}                 & \multicolumn{1}{l|}{} & \multicolumn{1}{l|}{0.916}                          & \multicolumn{1}{l|}{0.848}                          & \multicolumn{1}{l|}{0.732}                          & \multicolumn{1}{l|}{0.832}                       & \multicolumn{1}{l|}{} & \multicolumn{1}{l|}{0.354}                         & \multicolumn{1}{l|}{0.618}                         & \multicolumn{1}{l|}{0.486}                       & \multicolumn{1}{l|}{} & \multicolumn{1}{l|}{0.375}                          & \multicolumn{1}{l|}{0.156}                          & \multicolumn{1}{l|}{0.266}                       \\ \cline{1-1} \cline{3-6} \cline{8-10} \cline{12-14} 
\multicolumn{1}{|l|}{OC-SVM}                          & \multicolumn{1}{l|}{} & \multicolumn{1}{l|}{0.455}                          & \multicolumn{1}{l|}{0.655}                          & \multicolumn{1}{l|}{0.658}                          & \multicolumn{1}{l|}{0.589}                       & \multicolumn{1}{l|}{} & \multicolumn{1}{l|}{0.269}                         & \multicolumn{1}{l|}{0.635}                         & \multicolumn{1}{l|}{0.452}                       & \multicolumn{1}{l|}{} & \multicolumn{1}{l|}{0.508}                          & \multicolumn{1}{l|}{0.578}                          & \multicolumn{1}{l|}{0.543}                       \\ \cline{1-1} \cline{3-6} \cline{8-10} \cline{12-14} 
\multicolumn{1}{|l|}{SVDD}                            & \multicolumn{1}{l|}{} & \multicolumn{1}{l|}{\textbf{0.938}}                 & \multicolumn{1}{l|}{\textbf{0.906}}                 & \multicolumn{1}{l|}{0.839}                          & \multicolumn{1}{l|}{0.894}                       & \multicolumn{1}{l|}{} & \multicolumn{1}{l|}{0.305}                         & \multicolumn{1}{l|}{0.798}                         & \multicolumn{1}{l|}{0.552}                       & \multicolumn{1}{l|}{} & \multicolumn{1}{l|}{0.525}                          & \multicolumn{1}{l|}{0.658}                          & \multicolumn{1}{l|}{0.592}                       \\ \cline{1-1} \cline{3-6} \cline{8-10} \cline{12-14} 
\multicolumn{1}{|l|}{E-SVDD}                         & \multicolumn{1}{l|}{} & \multicolumn{1}{l|}{0.887}                          & \multicolumn{1}{l|}{0.835}                          & \multicolumn{1}{l|}{0.859}                          & \multicolumn{1}{l|}{0.860}                       & \multicolumn{1}{l|}{} & \multicolumn{1}{l|}{0.300}                         & \multicolumn{1}{l|}{0.000}                         & \multicolumn{1}{l|}{0.150}                       & \multicolumn{1}{l|}{} & \multicolumn{1}{l|}{0.000}                          & \multicolumn{1}{l|}{0.000}                          & \multicolumn{1}{l|}{0.000}                       \\ \cline{1-1} \cline{3-6} \cline{8-10} \cline{12-14} 
\multicolumn{1}{|l|}{GE-SVDD}                         & \multicolumn{1}{l|}{} & \multicolumn{1}{l|}{0.912}                          & \multicolumn{1}{l|}{0.883}                          & \multicolumn{1}{l|}{0.847}                          & \multicolumn{1}{l|}{0.881}                       & \multicolumn{1}{l|}{} & \multicolumn{1}{l|}{0.256}                         & \multicolumn{1}{l|}{0.814}                         & \multicolumn{1}{l|}{0.535}                       & \multicolumn{1}{l|}{} & \multicolumn{1}{l|}{\textbf{0.557}}                 & \multicolumn{1}{l|}{0.658}                          & \multicolumn{1}{l|}{\textbf{0.608}}              \\ \cline{1-1} \cline{3-6} \cline{8-10} \cline{12-14} 
\multicolumn{1}{|l|}{GE-SVM}                          & \multicolumn{1}{l|}{} & \multicolumn{1}{l|}{0.923}                          & \multicolumn{1}{l|}{0.899}                          & \multicolumn{1}{l|}{0.861}                          & \multicolumn{1}{l|}{\textbf{0.894}}              & \multicolumn{1}{l|}{} & \multicolumn{1}{l|}{0.385}                         & \multicolumn{1}{l|}{\textbf{0.907}}                & \multicolumn{1}{l|}{0.646}                       & \multicolumn{1}{l|}{} & \multicolumn{1}{l|}{0.542}                          & \multicolumn{1}{l|}{0.669}                          & \multicolumn{1}{l|}{0.605}                       \\ \cline{1-1} \cline{3-6} \cline{8-10} \cline{12-14} 
\end{tabular}\label{kernelresults}  
\end{table}

We further notice that by considering the class variance taking into account the concentration matrix in the regularization term, ES-SVDD performs better in most datasets as compared to the regularization terms without considering the concentration matrix. Overall $\Upsilon_2$ is found to be more robust that other regularization terms. Hence, we recommend to use samples belonging to the boundary and outside the boundary to describe the covariance of the class while taking into account the concentration matrix.

\begin{figure}[H]
	\centering
	\includegraphics[scale=0.40]{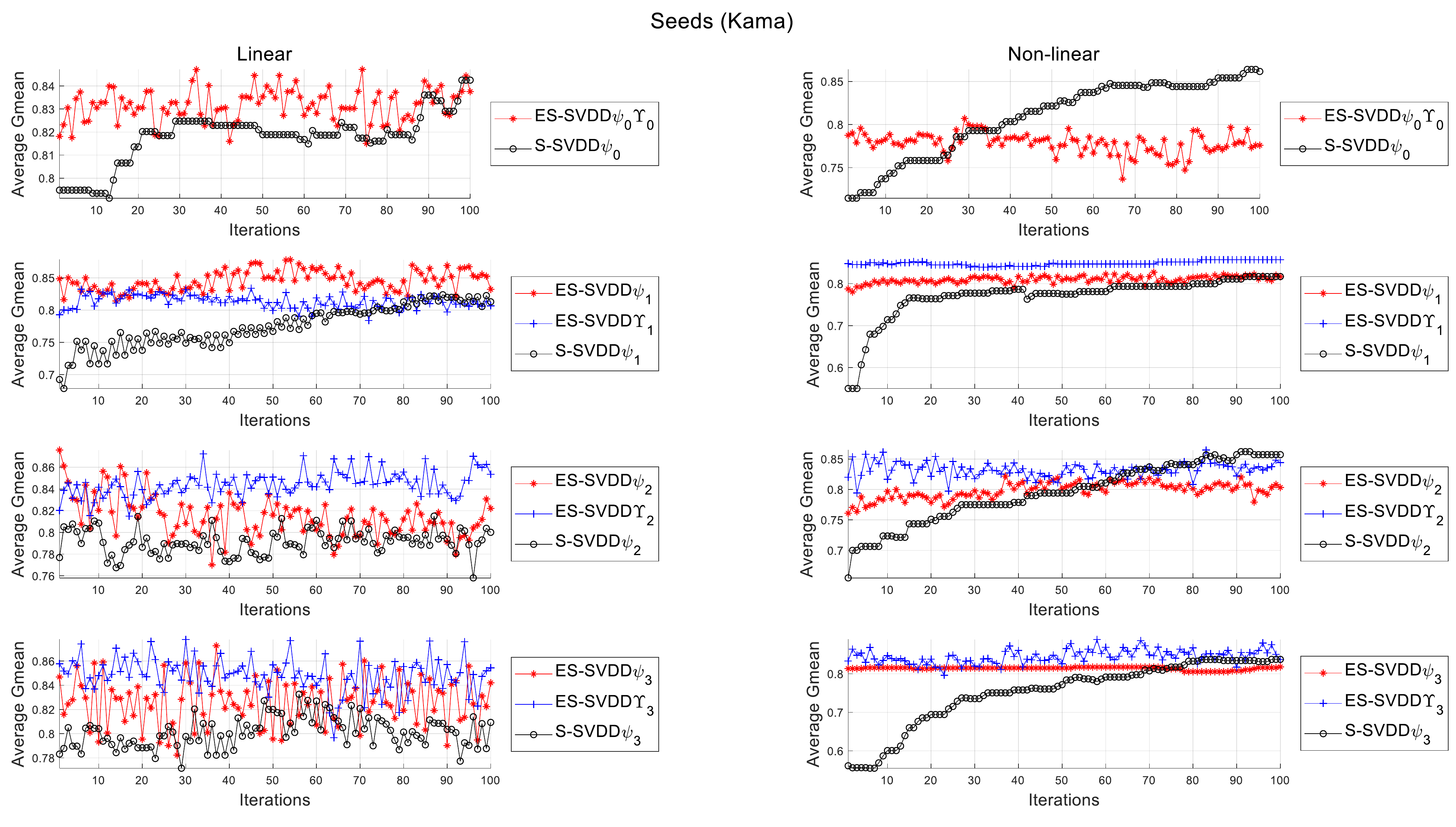}
		%\vspace{-2cm}
		%\rule{35em}{0.5pt}
	\caption{Comparison of different regularization terms for ES-SVDD and S-SVDD on dataset S-K}
	\label{unseeds}
\end{figure}

\begin{figure}[H]
	\centering
	\includegraphics[scale=0.40]{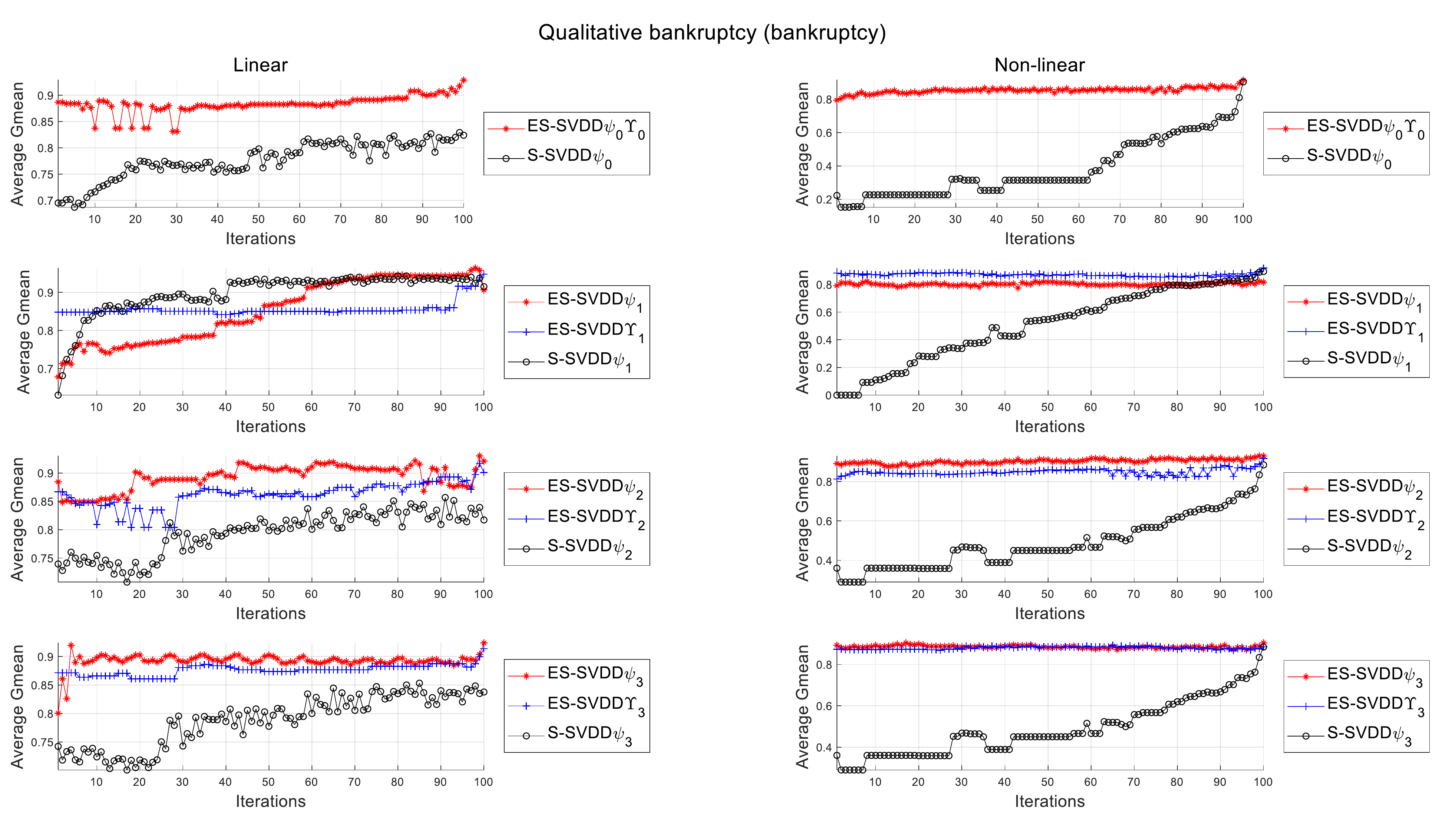}
		%\vspace{-2cm}
		%\rule{35em}{0.5pt}
	\caption{Comparison of different regularization terms for ES-SVDD and S-SVDD on dataset QB-B}
	\label{unbankruptcy}
\end{figure}

We also show the performance of the proposed ES-SVDD and the recently proposed S- SVDD on the test set after every training iteration for the linear and non-linear cases. We compare the performances of these methods with different regularization terms $\Upsilon$ and $\psi$. The average \textit{Gmean} value is calculated for each iteration over the 5 test splits for the different datasets, see Figures \ref{unseeds}-\ref{unbankruptcy} and Appendix.
It can clearly be seen from the figures that for both the linear and non-linear methods, ES-SVDD achieves its best performance much earlier than the recently proposed counterpart S-SVDD. This is not surprising, because the ellipsoidal description can fit a larger variety of data distributions, while the optimal spherical description gets successful only after the data variance for different dimensions has been equalized. Using the ellipsoidal data description in the proposed method makes it converge faster to an optimal solution. We also notice that for high dimensional datasets ES-SVDD$\psi_1$ and ES-SVDD$\Upsilon_1$ are more stable as compared to the other proposed linear and non-linear methods. Overall, the trend of a faster convergence and a higher stability in terms of producing consistent results for different range of iterations for ES-SVDD can be observed both in the linear and non-linear methods for all regularization terms in the majority of the cases.

\section{Conclusion}\label{conculsion}
In this paper, a novel method, ES-SVDD, for one-class classification is proposed. The proposed method projects the data from an input feature space to a new optimized subspace suitable for one-class classification. The proposed method generalizes S-SVDD for a hypersphere by using ellipsoidal data description. We proposed different regularization terms along with linear and non-linear formulations of the method. In most cases, the proposed ES-SVDD variants outperform the competing methods and converge faster than the case of data description without ellipsoidal encapsulation.

In the future, we intend to use other kernel types in the non-linear case of ES-SVDD. We also plan to devise a strategy for early exit in the training process to reduce the training time. We will also experiment with finetuning hyperparameters according to different criteria, such as area under receiver operating characteristic curve.
\section{Acknowledgement}\label{Acknowledgement}
This work was supported by a NSF-Business Finland CVDI project Amalia and a Business Finland projects VIRPA D and INDEX (DIMECC Industrial Data program).
\bibliographystyle{elsarticle-num}

\bibliography{ms}
\section*{Appendix}
\begin{figure}[H]
	\centering
	\includegraphics[scale=0.40]{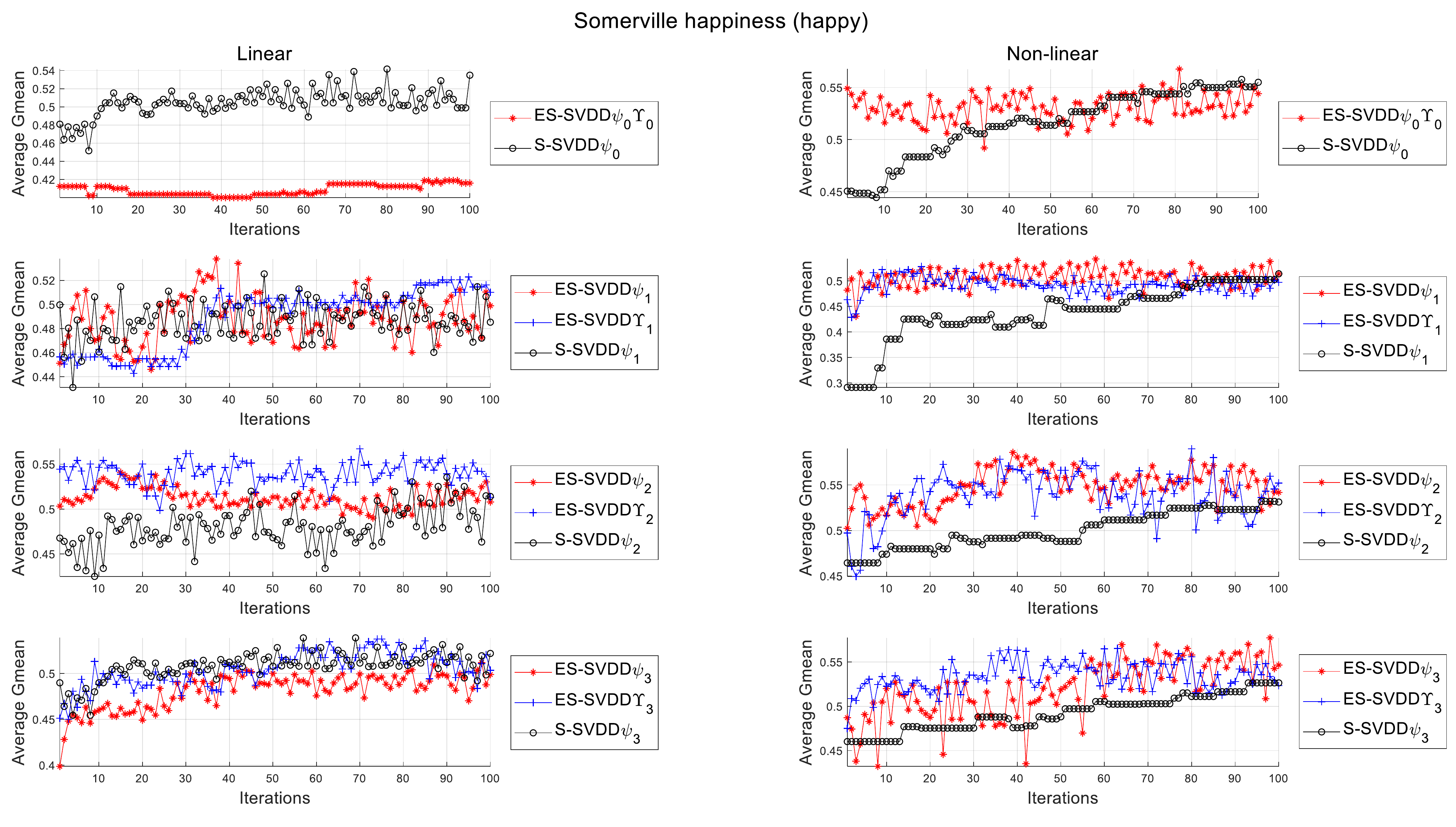}
		%\vspace{-2cm}
		%\rule{35em}{0.5pt}
	\caption{Comparison of different regularization terms for ES-SVDD and S-SVDD on dataset SH-H}
	\label{unhappiness}
\end{figure}
\begin{figure}[H]
	\centering
	\includegraphics[scale=0.40]{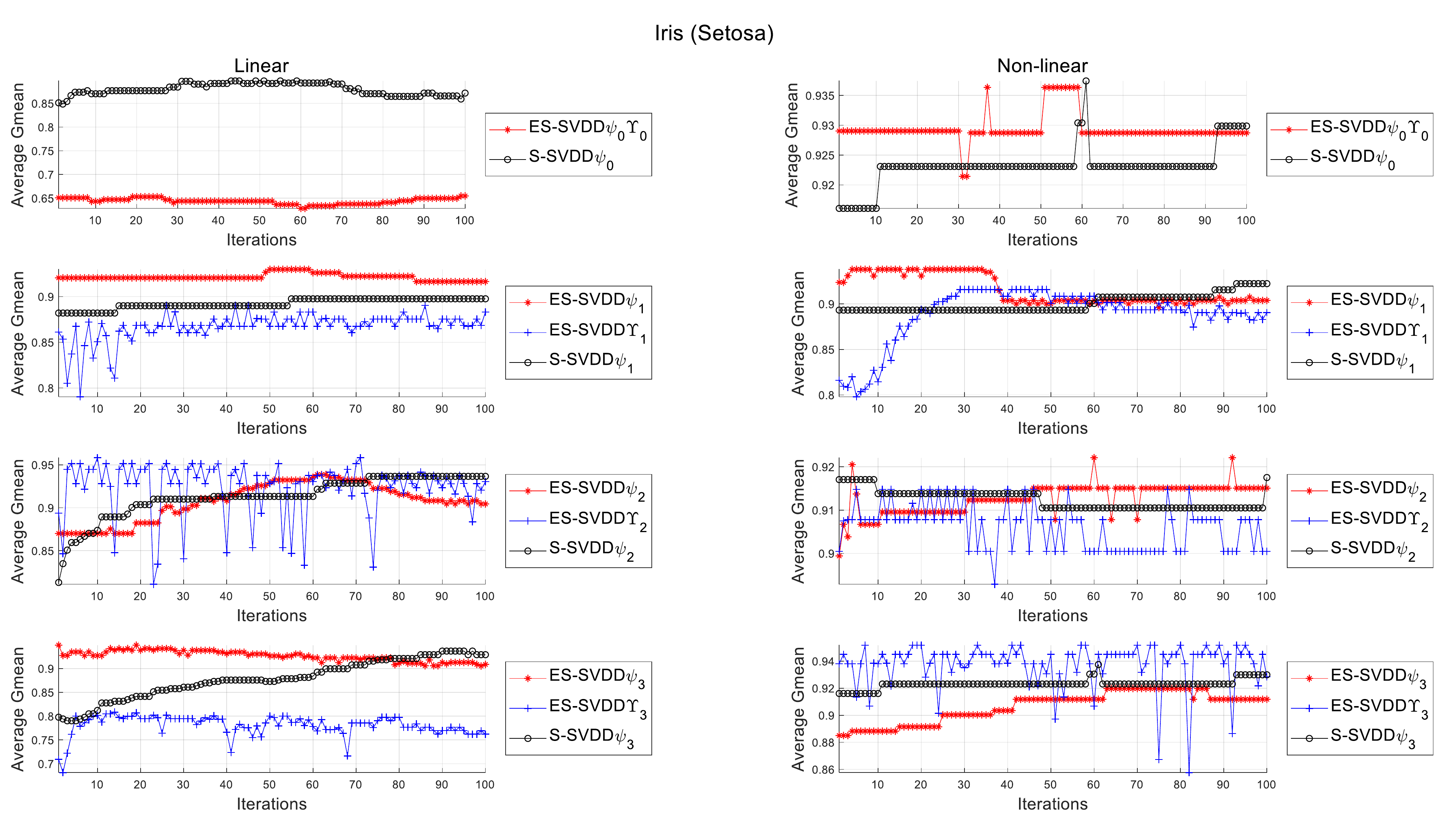}
		%\vspace{-2cm}
		%\rule{35em}{0.5pt}
	\caption{Comparison of different regularization terms for ES-SVDD and S-SVDD on dataset I-S}
	\label{uniris}
\end{figure}
\begin{figure}[H]
	\centering
	\includegraphics[scale=0.40]{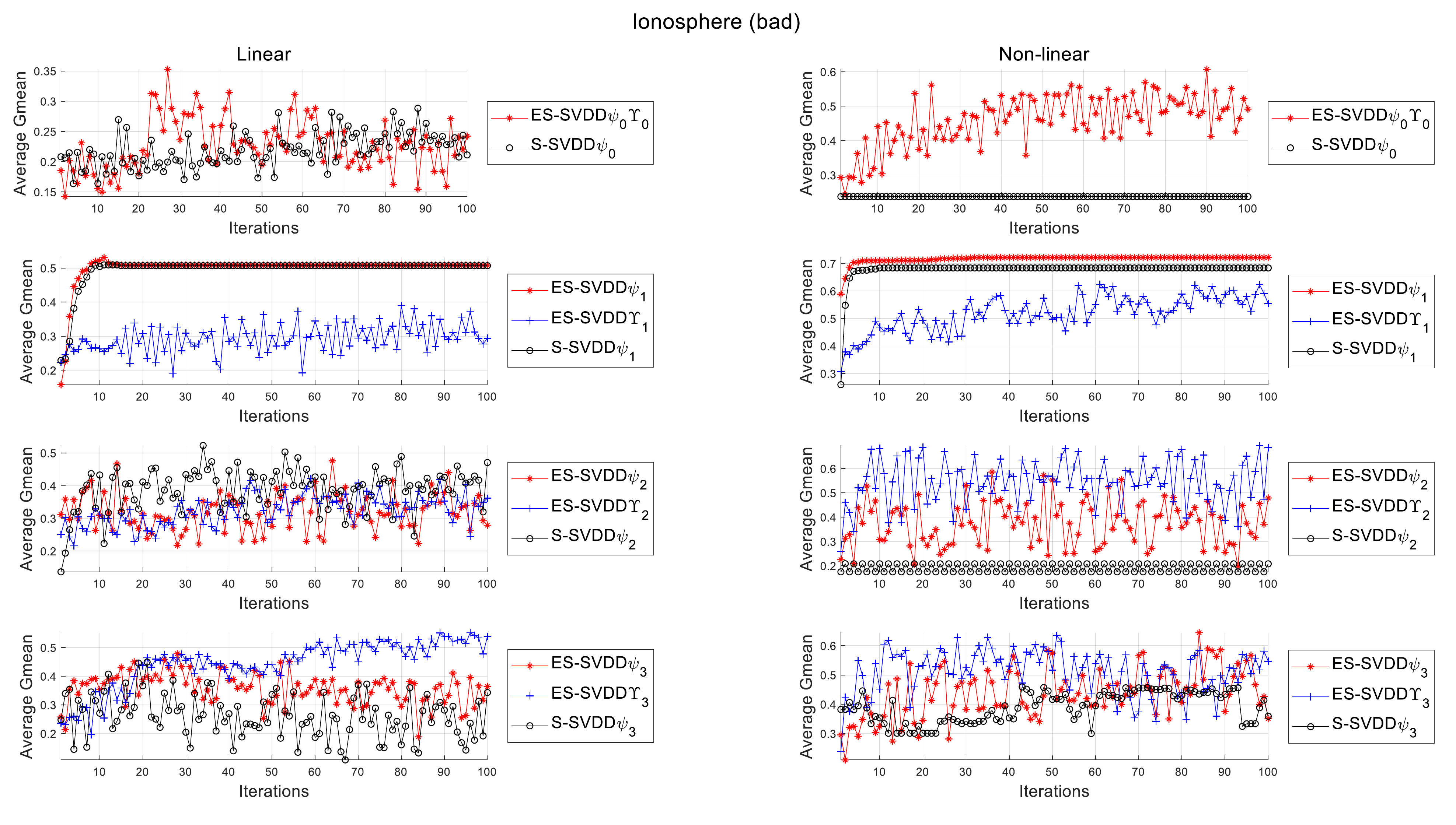}
		%\vspace{-2cm}
		%\rule{35em}{0.5pt}
	\caption{Comparison of different regularization terms for ES-SVDD and S-SVDD on dataset IS-B}
	\label{unionpsphere}
\end{figure}
\begin{figure}[H]
	\centering
	\includegraphics[scale=0.40]{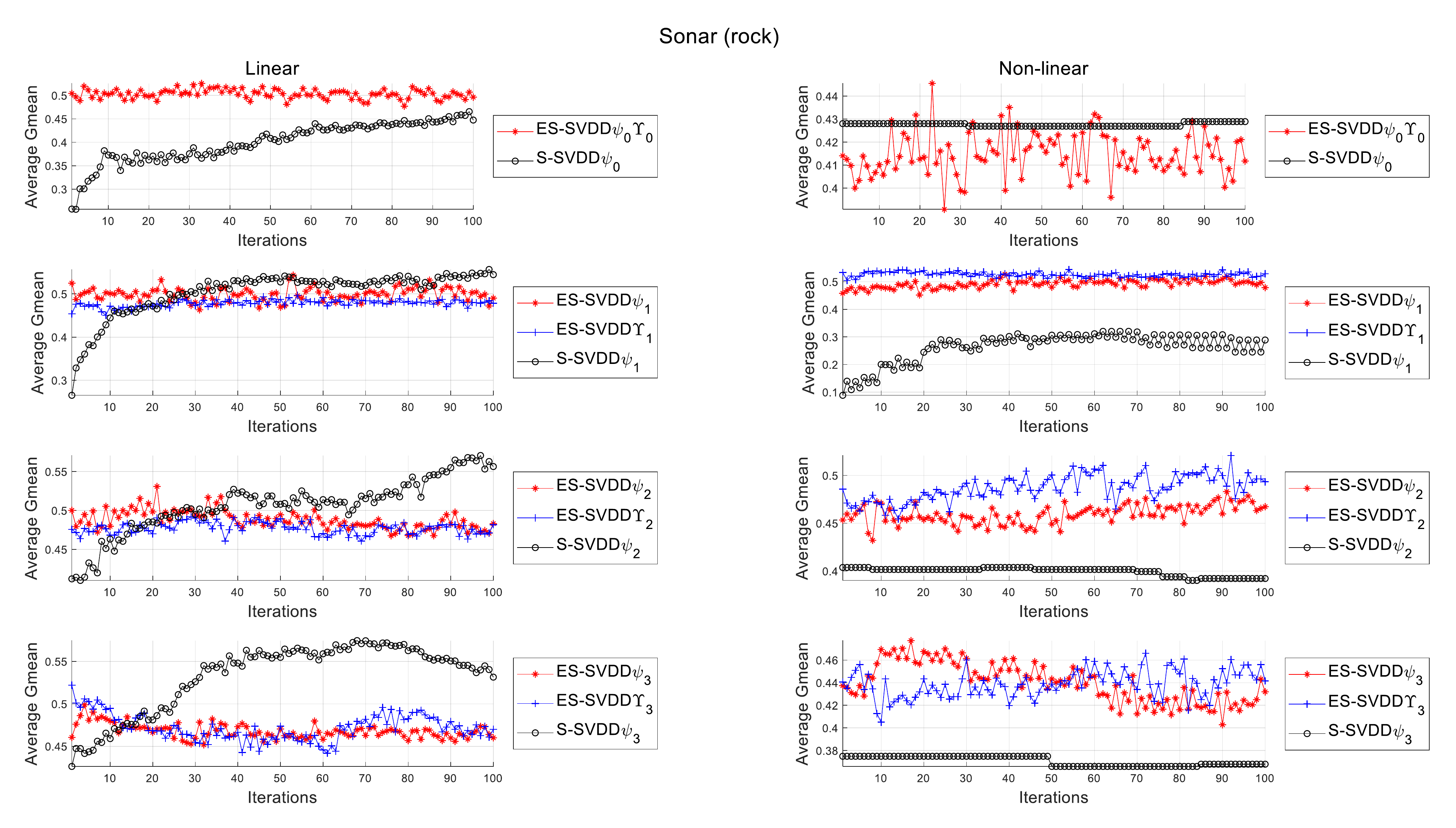}
		%\vspace{-2cm}
		%\rule{35em}{0.5pt}
	\caption{Comparison of different regularization terms for ES-SVDD and S-SVDD on dataset SR-R}
	\label{unsonar}
\end{figure}
\end{document}